\documentclass[10pt,journal,compsoc]{IEEEtran}
\ifCLASSINFOpdf
\else
\fi
\usepackage{times}
\usepackage{soul}
\usepackage{url}
\usepackage[hidelinks]{hyperref}
\usepackage[utf8]{inputenc}
\usepackage[small]{caption}
\usepackage{graphicx}
\usepackage{amsmath}
\usepackage{amsthm}
\usepackage{amssymb}
\usepackage{pifont}
\usepackage{booktabs}
\usepackage{algorithm}
\usepackage{algorithmic}
\usepackage{multirow}
\usepackage{array}
\usepackage{enumitem}
\usepackage[switch]{lineno}
\usepackage{color}
\usepackage{comment}
\DeclareMathOperator*{\argmin}{arg\,min}
\newcommand{\cmark}{\ding{51}}%
\newcommand{\xmark}{\ding{55}}%

\newcommand{\spara}[1]{\smallskip\noindent\textbf{#1}}
\newcolumntype{P}[1]{>{\centering\arraybackslash}p{#1}}
\usepackage{adjustbox}
\usepackage[figuresright]{rotating}
\usepackage{listings}
\usepackage{subfigure}

\begin{document}
%

\title{Automated Graph Machine Learning: Approaches, Libraries, \\Benchmarks and Directions}
%
%
%

\iffalse
	\author{
	First Author$^1$\footnote{Contact Author}\and
	Second Author$^2$\and
	Third Author$^{2,3}$\And
	Fourth Author$^4$\\
	\affiliations
	$^1$First Affiliation\\
	$^2$Second Affiliation\\
	$^3$Third Affiliation\\
	$^4$Fourth Affiliation\\
	\emails
	\{first, second\}@example.com,
	third@other.example.com,
	fourth@example.com
	}
\else
	\author{
	Ziwei Zhang\footnote{Equal contributions}   \and
	Xin Wang\footnotemark[1] \And
	Wenwu Zhu
	\affiliations
	Tsinghua University, Beijing, China\\
	\emails
	zw-zhang16@mails.tsinghua.edu.cn,
	\{xin\_wang,wwzhu\}@tsinghua.edu.cn
	}

\author{Xin Wang,~\IEEEmembership{Member,~IEEE,}
        Ziwei Zhang,~\IEEEmembership{Member,~IEEE,}
        Haoyang Li,~\IEEEmembership{Member,~IEEE,}
        and~Wenwu~Zhu,~\IEEEmembership{Fellow,~IEEE}
\IEEEcompsocitemizethanks{
\IEEEcompsocthanksitem Xin Wang, Ziwei Zhang, Haoyang Li and Wenwu Zhu are with the Department
of Computer Science and Technology, Tsinghua University, Beijing, China. 
Corresponding Authors: Wenwu Zhu.
E-mail: \{xin\_wang, zwzhang, wwzhu\}@tsinghua.edu.cn. lihy218@gmail.com
\IEEEcompsocthanksitem This work is supported by the National Key Research and Development Program of China No.2023YFF1205001, National Natural Science Foundation of China (No. 62222209, 62250008, 62102222). 
BNRist under Grant No. BNR2023RC01003, BNR2023TD03006. 
}
}

%
%

\fi

\markboth{Journal of \LaTeX\ Class Files,~Vol.~14, No.~8, August~2015}%
{Shell \MakeLowercase{\textit{et al.}}: Bare Demo of IEEEtran.cls for IEEE Journals}
%



\IEEEtitleabstractindextext{

\begin{abstract}
Graph machine learning has been extensively studied	in both academic and industry. However, as the literature on graph learning booms with a vast number of emerging methods and techniques, 
it becomes increasingly difficult to manually design the optimal machine learning algorithm for different graph-related tasks. 
To tackle the challenge, automated graph machine learning, which aims at discovering the best hyper-parameter and neural architecture configuration for different graph tasks/data without manual design, is gaining an increasing number of attentions 
from the research community. In this paper, we extensively discuss automated graph machine learning approaches, covering hyper-parameter optimization (HPO) and neural architecture search (NAS) for graph machine learning. 
We briefly overview existing libraries designed for either graph machine learning or automated machine learning respectively, and further in depth introduce AutoGL, our dedicated and the world's first open-source library for automated graph machine learning. Also, we describe a tailored benchmark that
supports unified, reproducible, and efficient evaluations.
Last but not least, we share our insights on future research directions for automated graph machine learning. 
This paper is the first systematic and comprehensive discussion of approaches, libraries as well as directions for automated graph machine learning.
\end{abstract}

\begin{IEEEkeywords}
Graph Machine Learning, Graph Neural Network, Automated Machine Learning, AutoML, Neural Architecture Search, Hyper-parameter Optimization
\end{IEEEkeywords}
}

%
\maketitle
\IEEEdisplaynontitleabstractindextext
\IEEEpeerreviewmaketitle

\IEEEraisesectionheading{\section{Introduction}}
\IEEEPARstart{G}{raph} data is ubiquitous in our daily life. We can use graphs to model the complex relationships and dependencies between entities ranging from small molecules in proteins and particles in physical simulations to large national-wide power grids and global airlines. Therefore, graph machine learning, i.e., machine learning on graphs, has long been an important research direction for both academics and industry~\cite{newman2018networks}. In particular, network embedding~\cite{cui2018survey,hamilton2017representation,goyal2018graph,cai2018comprehensive} and graph neural networks (GNNs)~\cite{zhang2020deep,wu2020comprehensive,zhou2018graph} have drawn increasing attention in the last decade. They are successfully applied to recommendation systems~\cite{ying2018graph,ma2019learning,li2021intention,liu2023graph}, 
information retrieval~\cite{cui2022can,feng2022dc,wang2022gnn,liang2023abslearn},
fraud detection~\cite{akoglu2015graph}, bioinformatics~\cite{su2020network,zitnik2017predicting}, physical simulation~\cite{kipf2018neural}, traffic forecasting~\cite{li2018dcrnn_traffic,yu2018spatio}, knowledge representation~\cite{wang2017knowledge}, drug re-purposing~\cite{ioannidis2020few,gysi2020network} and pandemic prediction~\cite{kapoor2020examining} for Covid-19.

Despite the popularity of graph machine learning algorithms, the existing literature heavily relies on manual hyper-parameter or architecture design
to achieve the best performance,
resulting in costly human efforts when a vast number of models emerge for various graph tasks. 
Take GNNs as an example, at least one hundred new general-purpose architectures have been published in top-tier machine learning and data mining conferences in the year of 2021 alone, not to mention cross-disciplinary researches of task-specific designs. More and more human efforts are inevitably needed if we stick to the manual try-and-error paradigm in designing the optimal algorithms for targeted tasks.

On the other hand, automated machine learning (AutoML) has been extensively studied to reduce human efforts in developing and deploying machine learning models~\cite{he2020automl,yao2018taking}. Complete AutoML pipelines have the potential to automate every step of machine learning, including auto data collection and cleaning, auto feature engineering, and auto model selection and optimization, etc. Due to the popularity of deep learning models, hyper-parameter optimization (HPO)~\cite{bergstra2012random,bergstra2011algorithms,snoek2012practical,liu2021meta} and neural architecture search (NAS)~\cite{elsken2019neural,wei2021autoias} are most widely studied. AutoML has achieved or surpassed human-level performance~\cite{zoph2017neural,liu2018darts,pham2018efficient} with little human guidance in areas such as computer vision~\cite{zoph2018learning,real2019regularized}.

Automated graph machine learning, combining advantages of AutoML and graph machine learning, naturally serves as a promising research direction to further boost the model performance, which has attracted an increasing number of interests from the community.
In this paper, we provide a systematic overview of approaches for automated graph machine learning\footnote{We provide a paper collection about automated graph machine learning at \url{https://github.com/THUMNLab/awesome-auto-graph-learning}.}, introduce related public libraries as well as our AutoGL, the world's first open-source library for automated graph machine learning, describe a tailored benchmark that supports unified, reproducible, and efficient evaluations,
and share our insights on challenges and future research directions. 

Particularly, we focus on two major topics: HPO and NAS of graph machine learning. For HPO, we focus on how to develop scalable methods. For NAS, we follow the literature and compare different methods from search spaces, search strategies, and performance estimation strategies. We also briefly discuss several recent automated graph learning works that feature in different aspects such as architecture pooling, structure learning, accelerator and joint software-hardware design etc. Besides, how different methods tackle the challenges of AutoML on graphs are discussed along the way as well. 
Then, we review libraries related to automated graph machine learning and discuss AutoGL, the first dedicated framework and open-source library for automated graph machine learning. 
We highlight the design principles of AutoGL and briefly introduce its usages, which are all specially designed for AutoML on graphs. 
Last but not least, we point out the potential research directions for both graph HPO and graph NAS, including but not limited to \textit{Scalability, Explainability, Out-of-distribution generalization, Robustness}, and \textit{Hardware-aware design} etc. 
We believe this paper will greatly facilitate and further promote the studies and applications of automated graph machine learning in both academia and industry.

The rest of the paper is organized as follows. In Section~\ref{sec:review}, we intoduce the fundamentals and preliminaries for automated graph machine learning by briefly introducing basic formulations of graph machine learning and AutoML. We comprehensively discuss HPO based approaches on graph machine learning in Section~\ref{sec:hpo} and NAS based methods for graph machine learning in Section~\ref{sec:nas}. Then, in Section~\ref{sec:tool}, we overview related libraries for graph machine learning and automated machine learning and in depth introduce AutoGL, our dedicated and the world's first open-source library tailored for automated graph machine learning. 
We discuss the tailored benchmark that enables fair, fully reproducible, and efficient empirical comparisons in Section~\ref{sec:benchmark}.
Last but not least, we outline future research opportunities in Section~\ref{sec:future} and conclude the whole paper in Section~\ref{sec:conclusion}. 

\section{Fundamentals and Preliminaries of Automated Graph Machine Learning}\label{sec:review}

We briefly present basic problem formulations for graph machine learning, automated machine learning as well as unique characteristics for automated graph machine learning before moving to the next section.

\subsection{Graph Machine Learning}\label{sec:mlgform}
Consider a graph $\mathcal{G} = \left( \mathcal{V},\mathcal{E}\right)$ where $\mathcal{V} = \left\{v_1,v_2,...,v_{\left|\mathcal{V}\right|}\right\}$ is a set of nodes and $\mathcal{E} \subseteq \mathcal{V} \times \mathcal{V}$ is a set of edges. The neighborhood of node $v_i$ is denoted as $\mathcal{N}(i)=  \left\{v_j: (v_i,v_j) \in \mathcal{E}\right\}$. The nodes can also have features denoted as $\mathbf{F}\in \mathbb{R}^{\left|\mathcal{V} \right| \times f}$, where $f$ is the number of features. We use bold uppercases (e.g., $\mathbf{X}$) and bold lowercases (e.g., $\mathbf{x}$) to represent matrices and vectors, respectively.

Most tasks of graph machine learning can be divided into the following two categories: 
\begin{itemize}
	\item Node-level tasks: the tasks are associated with individual nodes or pairs of nodes. Typical examples include node classification and link prediction. 
	\item Graph-level tasks: the tasks are associated with the whole graph, such as graph classification and graph generation.	
\end{itemize}
For node-level tasks, graph machine learning models usually learn a node representation $\mathbf{H} \in \mathbb{R}^{\left|\mathcal{V}\right| \times d}$ and then adopt a classifier or predictor on the node representation to solve the task. For graph-level tasks, a representation for the whole graph is learned and fed into a classifier/predictor.

GNNs are the current state-of-the-art in learning node and graph representations.
The message-passing framework of GNNs~\cite{gilmer2017neural} is formulated as follows.
\begin{gather}\label{eq:mpnn}
    \mathbf{m}^{(l)}_i = \text{AGG}^{(l)}\left(\left\{a_{ij}^{(l)} \mathbf{W}^{(l)} \mathbf{h}^{(l)}_i, \forall j \in \mathcal{N}(i) \right\} \right) \\
    \mathbf{h}^{(l+1)}_i = \sigma\left(\text{COMBINE}^{(l)}\left[ \mathbf{m}^{(l)}_i, \mathbf{h}^{(l)}_i\right] \right),
\end{gather}
where $\mathbf{h}^{(l)}_i$ denotes the node representation of node $v_i$ in the $l^{th}$ layer, $\mathbf{m}^{(l)}$ is the message for node $v_i$, $\text{AGG}^{(l)}(\cdot)$ is the aggregation function, $a_{ij}^{(l)}$ denotes the weights from node $v_j$ to node $v_i$, $\text{COMBINE}^{(l)}(\cdot)$ is the combining function, $\mathbf{W}^{(l)}$ are learnable weights, and $\sigma(\cdot)$ is an activation function. The node representation is usually initialized as node features $\mathbf{H}^{(0)} = \mathbf{F}$, and the final representation is obtained after $L$ message-passing layers $\mathbf{H} = \mathbf{H}^{(L)}$.

For the graph-level representation, pooling methods (also called readout) are applied to the node representations
\begin{equation}\label{eq:pool}
	\mathbf{h}_{\mathcal{G}} = \text{POOL}\left(\mathbf{H}\right),
\end{equation}
i.e., $\mathbf{h}_{\mathcal{G}}$ is the representation of $\mathcal{G}$.

\subsection{Automated Machine Learning (AutoML)}\label{sec:automlform}
Many AutoML algorithms such as HPO and NAS can be formulated as the following bi-level optimization problem:
\begin{equation}\label{eq:nasobj}
\begin{split}
	 & \min_{\alpha \in \mathcal{A}} \mathcal{L}_{val}\left(\mathbf{W}^*(\alpha),\alpha\right) \\
      \text{s.t.} \quad & \mathbf{W}^*(\alpha) =  \argmin_{\mathbf{W}} \left(\mathcal{L}_{train}\left(\mathbf{W},\alpha\right)\right) ,
\end{split}
\end{equation}
where $\alpha$ is the optimization objective of the AutoML algorithm, e.g., hyper-parameters in HPO and neural architectures in NAS, $\mathcal{A}$ is the feasible space for the objective, and $\mathbf{W}(\alpha)$ are trainable weights in the graph machine learning models. Essentially, we aim to optimize the objective in the feasible space so that the model achieves the best results in terms of a validation function, and $\mathbf{W}^*$ indicates that the weights are fully optimized in terms of a training function. Different AutoML methods differ in how the feasible space is designed and how the objective functions are instantiated and optimized since directly optimizing Eq.~\eqref{eq:nasobj} requires enumerating and training every feasible objective, which is prohibitive in practice.

Typical formulations of automated graph machine learning need to properly integrate the above formulations in Section~\ref{sec:mlgform} and Section~\ref{sec:automlform} to form a new optimization problem. 

\subsection{Automated Graph Machine Learning}\label{sec:autographform}
Automated graph machine learning, which non-trivially combines the strength of AutoML and graph machine learning, faces the following challenges.
\begin{itemize}
\item \textbf{The uniqueness of graph machine learning:} Unlike audio, image, or text, which has a grid structure, graph data lies in a non-Euclidean space~\cite{bronstein2017geometric}. Thus, graph machine learning usually has unique architectures and designs. For example, typical NAS methods focus on the search space for convolution and recurrent operations, which is distinct from the building blocks of GNNs~\cite{ijcai2020-195}.
\item \textbf{Complexity and diversity of graph tasks:} As aforementioned, graph tasks per se are complex and diverse, ranging from node-level to graph-level problems, and with different settings, objectives, and constraints~\cite{hu2020open}. How to impose proper \emph{inductive bias} and integrate \emph{domain knowledge} into a graph AutoML method is indispensable.
\item \textbf{Scalability:} Many real graphs such as social networks or the Web are incredibly large-scale with billions of nodes and edges~\cite{zang2018power}. Besides, the nodes in the graph are interconnected and cannot be treated as independent samples. Designing scalable AutoML algorithms for graphs poses significant challenges since both graph machine learning and AutoML are already notorious for being compute-intensive.
\end{itemize}
Approaches with HPO or NAS for graph machine learning reviewed in later sections target at handling at least one of these three challenges. 
As such, we will discuss approaches for automated graph machine learning from two aspects: i) HPO for graph machine learning and ii) NAS for graph machine learning.

\section{HPO for Graph Machine Learning}\label{sec:hpo}
In this section, we review HPO for graph machine learning. The main challenge here is scalability, i.e., a real graph can have billions of nodes and edges, and each trial on the graph is computationally expensive. Next, we elaborate on how different methods tackle the efficiency challenge. Notice that we omit some straightforward HPO methods such as random search and grid search~\cite{bergstra2012random} since they are applied to graphs without any modification.

Tu~\textit{et al.}~\cite{tu2019autone} propose AutoNE, the first HPO method specially designed to tackle the efficiency problem of graphs, to facilitate the graph hyper-parameter optimization for large-scale graph representation learning. AutoNE proposes a transfer paradigm that samples subgraphs as proxies for the large graph. Specifically, AutoNE has three modules: the sampling module, the signature extraction module, and the meta-learning module. In the sampling module, multiple representative subgraphs are sampled from the large graph using a multi-start random walk strategy. Each subgraph learns a representation by the signature extraction module. Then, AutoNE conducts HPO on the sampled subgraphs using Bayesian optimization~\cite{snoek2012practical} and records the results. Finally, using the HPO results and representation of subgraphs to extract meta-knowledges, AutoNE fine-tunes hyper-parameters on the large graph using the meta-learning module. In this way, AutoNE achieves satisfactory results while maintaining scalability since the knowledge of multiple HPO trials on the sampled subgraphs and a few HPO trails on the large graph are properly integrated.

\textcolor{black}{Wang~\textit{et al.}~\cite{wang2021explainable} propose e-AutoGR to further increase the explainability of hyper-parameter optimization for automated graph representation learning, with the help of hyper-parameter importance decorrelation. e-AutoGR employs six fully 
explainable graph features, i.e., \textit{number of nodes, number of edges, number of triangles, global clustering coefficient, maximum total degree value} and \textit{number of components}, as measures for similarity between different graphs. A hyper-parameter decorrelation algorithm (HyperDeco) is proposed to decorrelate the mixed relations among different hyper-parameters given various graph features so that more accurate importance of different hyper-parameters towards model performances can be estimated through any regression approaches.
The authors theoretically validate the correctness of the proposed hyper-parameter decorrelation algorithm and empirically discover that \textit{first-order proximity} is most important for AROPE~\cite{zhang2018arbitrary}, \textit{number of walks} together with \textit{window size} is of great importance for DeepWalk~\cite{perozzi2014deepwalk}, and \textit{dropout} is particularly important for GCN~\cite{kipf2017semi}.}

Guo~\textit{et al.}~\cite{guo2021jitune} propose ITuNE to replace the sampling process of AutoNE with graph coarsening to generate a hierarchical graph synopsis. A similarity measurement module is also proposed to ensure that the coarsened graph shares sufficient similarity with the large graph. Compared with sampling, such graph synopsis can better preserve graph structural information. Therefore, JITuNE argues that the best hyper-parameters in the graph synopsis can be directly transferred to the large graph. Besides, since the graph synopsis is generated in a hierarchy, the granularity can be more easily adjusted to meet the time constraints of downstream tasks. 

Yuan~\textit{et al.}~\cite{yuan2021novel} propose HESGA as another strategy to improve efficiency using a hierarchical evaluation strategy together with evolutionary algorithms. Specifically, HESGA proposes to evaluate the potential of hyper-parameters by interrupting training after a few epochs and calculating the performance gap with respect to the initial performance with random model weights. This gap is used as a fast score to filter out unpromising hyper-parameters. Then, the standard full evaluation, i.e., training until convergence, is adopted as the final assessor to select the best hyper-parameters to be stored in the population of the evolutionary algorithm.

Besides efficiency, Yoon~\textit{et al.}~\cite{yoon2020autonomous} propose AutoGM to focuses on studying a unified framework for various graph machine learning algorithms. Specifically, AutoGM finds that many popular GNNs and PageRank can be characterized in a framework similar to Eq.~\eqref{eq:mpnn} with five hyper-parameters: the number of message-passing neighbors, the number of message-passing steps, the aggregation function, the dimensionality, and the non-linearity. AutoGM also adopts Bayesian optimization to optimize these hyper-parameters.

\textcolor{black}{Yuan~\textit{et al.}~\cite{yuan2021hyperparameters} focus on the impact of selecting two types of GNN hyper-parameters (i.e., graph-related layers and task-specific layers) on the performance of GNN for molecular property prediction. They employed CMA-ES for HPO, which is a derivative-free and evolutionary black-box optimization method. The results reveal that optimizing the two types of hyper-parameters separately can result in improvement on GNN performance, and removing any of the two types of hyper-parameters may result in deteriorated performance. Even doing this means a larger search space, which seems to be more challenging given the same number of trials (limited computational resources), such a strategy can surprisingly achieve better performance. Meanwhile, their study further confirms the importance of HPO for GNNs in molecular property prediction problems.}

\textcolor{black}{Many molecular datasets are far smaller than other datasets in typical deep learning applications. Most HPO methods have not been explored in terms of their performances on these small datasets in molecular domain. Yuan et~al.~\cite{yuan2021systematic} conduct a theoretical analysis of common and specific features for two state-of-the-art HPO algorithms: i.e., TPE and CMA-ES, and they compare them with random search (RS). Experimental studies are carried out on several benchmarks in MoleculeNet, from different perspectives to investigate the impact of RS, TPE, and CMA-ES on HPO of GNNs for molecular property prediction. Their experimental results indicate that TPE is the most suited HPO method for GNN under molecular property prediction problems with limited computational resources. Meanwhile, RS is the simplest method capable of achieving comparable performance with TPE and CMA-ES.}

\textcolor{black}{GCN models are sensitive to the choice of hyper-parameters such as dropout rate and learning weight decay~\cite{wang2021explainable}, especially for deep GCN models. Zhu et~al.~\cite{zhu2021automated} therefore target at automating the training of GCN models through hyper-parameter optimization. To be specific, they propose a self-tuning GCN (ST-GCN) approach by incorporating \textit{hypernets} in each graph convolutional layer, enabling the joint optimization over GCN model parameters and hyper-parameters. They further extend the approach through incorporating the population based training scheme and adopt a population based training framework to self-tuning GCN, thus alleviating local minima problem via exploring hyper-parameter space globally. Experimental results on three benchmark datasets demonstrate the effectiveness of their approaches in terms of optimizing multi-layer GCNs.}

\textcolor{black}{Bu et~al.~\cite{bu2021automatic} analyze the performance of different evolutionary algorithms on automated graph machine learning through experimental study. The experimental results show that evolutionary algorithms can serve as an effective alternative to the traditional hyper-parameter optimization algorithms such as random search, grid search and Bayesian Optimization for GNN.}

\textcolor{black}{Sun et~al.~\cite{sun2021automated} propose AutoGRL, an automated graph representation learning framework for node classification task. AutoGRL consists of an appropriate search space with four components: data augmentation, feature engineering, hyper-parameter optimization, and architecture search. Given graph data, AutoGRL searches for the best graph representation learning model in the search space using an efficient searching algorithm. Extensive experiments are conducted on four real-world node classification datasets to demonstrate that AutoGRL can automatically find competitive graph representation learning models on specific graph data effectively and efficiently.}

\textcolor{black}{Yang et~al.~\cite{yang2022calibrate}} address the underexplored issue of obtaining reliable and trustworthy predictions using automated Graph Neural Networks (GNNs). It integrates uncertainty estimation into the Hyperparameter Optimization (HPO) problem through a bilevel formulation in a novel model named HyperU-GCN. The upper-level problem focuses on reasoning uncertainties by developing a probabilistic hypernetwork through a variational Bayesian approach. The lower-level problem targets how the weights in the Graph Convolutional Network (GCN) respond to a distribution of hyperparameters. By incorporating model uncertainty into the hyperparameter space, HyperU-GCN is able to achieve calibrated predictions, similar to Bayesian model averaging over hyperparameters. Experimental results on six public datasets indicate that this approach outperforms several state-of-the-art methods in terms of node classification accuracy and expected calibration error (ECE).

\textcolor{black}{Lloyd et~al.~\cite{lloyd2023assessing}} focus on the challenges of embedding knowledge graphs into low-dimensional spaces, a process that is computationally expensive largely due to hyperparameter optimization. They introduce a novel approach using Sobol sensitivity analysis to evaluate the significance of different hyperparameters in affecting the quality of the embeddings. Through thousands of trials and subsequent regression analysis, they identify considerable variability in the importance of different hyperparameters across various knowledge graphs. This variability is attributed to differences in dataset characteristics. Additionally, the paper makes a unique contribution by identifying data leakage issues in the UMLS knowledge graph and presenting a leakage-robust variant, termed UMLS-43.

\textcolor{black}{Yoon et~al.~\cite{yoon2022autonomous}} address the challenge of selecting the most suitable graph algorithm for specific real-world applications due to the proliferation of algorithms with different problem formulations, computational times, and memory footprints. To resolve this, they propose AUTOGM, an automated system for graph mining algorithm development. The paper introduces a unified framework, UNIFIEDGM, which simplifies the search space for graph algorithms by requiring only five parameters for algorithm determination. AUTOGM then uses Bayesian Optimization to find the optimal parameter set for UNIFIEDGM. To assess algorithmic efficacy within a given computational budget, the authors introduce a novel budget-aware objective function. Tests on various real-world datasets show that AUTOGM generates novel graph algorithms that offer the best speed-accuracy trade-off compared to existing models.

\textcolor{black}{Zhang et~al.~\cite{zhang2022kgtuner}} address the issue of inefficient hyper-parameter (HP) tuning in the context of knowledge graph (KG) learning. The authors first conduct a thorough analysis of different hyper-parameters and their transferability from smaller subgraphs to full graphs. Based on these insights, they introduce a two-stage search algorithm called KGTuner. In the first stage, the algorithm efficiently explores hyper-parameter configurations using small subgraphs. In the second stage, the best-performing configurations are fine-tuned on the full, large-scale graph. Experimental results demonstrate that KGTuner outperforms baseline algorithms, achieving an average relative improvement of 9.1\% across four different embedding models when applied to large-scale KGs in the open graph benchmark.

\textcolor{black}{Yang et~al.~\cite{yang2022revisiting}} present a systematic analysis of the impact of hyperparameters on both factorization-based and graph-sampling-based graph embedding techniques for homogeneous graphs. The authors design generalized techniques that include a wide range of hyperparameters and conduct an exhaustive experimental study with over 3,000 trained embedding models per dataset. The findings reveal that optimal hyperparameter settings, rather than the complexity of the embedding models, largely account for performance gains. The study shows that well-tuned hyperparameters can outperform a collection of 18 state-of-the-art graph embedding models by a margin of 0.30-35.41\% across various tasks. Importantly, the paper notes that there is no universal set of hyperparameters that are optimal for all tasks, but offers task-specific recommendations for hyperparameter settings, which can serve as valuable guidelines for future research in embedding-based graph analyses.

\section{NAS for Graph Machine Learning}\label{sec:nas}
  
  \begin{table*}[htbp]
	\centering
    \caption{A summary of different graph neural architecture search (NAS) methods for automated graph machine learning.}
    \small
    \begin{adjustbox}{angle=90}

		\begin{tabular}{c|P{0.5cm}P{0.55cm}P{0.75cm}P{0.35cm}P{0.8cm}|P{0.45cm}P{0.65cm}|c|c|c} \toprule
                \multirow{2}{*}{Method}  & \multicolumn{5}{c|}{Search space} & \multicolumn{2}{c|}{Tasks} & \multirow{2}{*}{Search Strategy} & Performance  &\multirow{2}{*}{Other Characteristics}  \\
			                                        & Micro & Macro &Pooling& HP    & Layers & Node   & Graph &                               &Estimation &- \\ \midrule 
			GraphNAS~\cite{ijcai2020-195}           &\cmark &\cmark &\xmark &\xmark & Fixed  & \cmark & \xmark& RNN controller + RL           & -         &- \\ 
			AGNN~\cite{zhou2019auto}                &\cmark &\xmark &\xmark &\xmark & Fixed  & \cmark & \xmark& Self-designed controller + RL & Inherit Weights  &- \\ 
			SNAG~\cite{zhao2020simplifying}         &\cmark &\cmark &\xmark &\xmark & Fixed  & \cmark & \xmark& RNN controller + RL           & Inherit Weights  &Simplify the micro search space \\ 
			PDNAS~\cite{zhao2020probabilistic}      &\cmark &\cmark &\xmark &\xmark & Fixed  & \cmark & \xmark& Differentiable                & One-shot  &- \\ 
			NAS-GNN~\cite{nunes2020neural}          &\cmark &\xmark &\xmark &\cmark & Fixed  & \cmark & \xmark& Evolutionary algorithm        & -         &- \\ 
			AutoGraph~\cite{li2020autograph}        &\cmark &\cmark &\xmark &\cmark & Various& \cmark & \xmark& Evolutionary algorithm        & -         &- \\ 
			GeneticGNN~\cite{shi2020evolutionary}   &\cmark &\xmark &\xmark &\cmark & Fixed  & \cmark & \xmark& Evolutionary algorithm        & -         &- \\ 
			EGAN~\cite{zhao2021efficient}           &\cmark &\cmark &\xmark &\xmark & Fixed  & \cmark & \cmark& Differentiable                & One-shot  & Sample small graphs for efficiency \\ 
			NAS-GCN~\cite{jiang2020graph}           &\cmark &\cmark &\cmark &\xmark & Fixed  & \xmark & \cmark& Evolutionary algorithm        & -         & Handle edge features               \\ 
			LPGNAS~\cite{zhao2020learned}           &\cmark &\cmark &\xmark &\xmark & Fixed  & \cmark & \xmark& Differentiable                & One-shot  & Search for quantization options    \\ 
			GraphGym~\cite{you2020design}            &\cmark &\cmark &\xmark &\cmark & Various& \cmark & \cmark & Random search                & -         & Transfer across datasets and tasks \\ 
			SGAS~\cite{li2020sgas}                  &\cmark &\xmark &\xmark &\xmark & Fixed  & \cmark & \cmark & Self-designed algorithm      & -         &-    \\ 
		Peng~\textit{et al.}~\cite{peng2020learning}&\cmark &\xmark &\xmark &\xmark & Fixed  & \xmark & \cmark &CEM-RL~\cite{pourchot2018cemrl}&- & Search spatial-temporal modules \\
			GNAS~\cite{cai2021rethinking}           &\cmark &\cmark &\xmark &\xmark & Various& \cmark & \cmark & Differentiable & One-shot & - \\
			AutoSTG~\cite{pan2021autostg}           &\xmark &\cmark &\xmark &\xmark & Fixed  & \cmark & \xmark & Differentiable    & One-shot+meta learning & Search spatial-temporal modules \\
			DSS~\cite{li2021one}                    &\cmark &\cmark &\xmark &\xmark & Fixed  & \cmark & \xmark & Differentiable & One-shot & Dynamically update search space  \\
			SANE~\cite{zhao2021search}              &\cmark &\cmark &\xmark &\xmark & Fixed  & \cmark & \xmark & Differentiable & One-shot & - \\	
			AutoAttend~\cite{guan2021autoattend}    &\cmark &\cmark &\xmark &\xmark & Fixed  & \cmark & \cmark & Evolutionary algorithm & One-shot & Cross-layer attention \\
			DiffMG~\cite{ding2020propagation}       &\cmark &\cmark &\xmark &\xmark & Fixed  & \cmark & \xmark & Differentiable                & One-shot  & Support heterogeneous graphs \\ 
			DeepGNAS~\cite{feng2021search}          &\cmark &\cmark &\xmark &\xmark & Various& \cmark & \xmark & Controller +RL                & -         & Alleviate over-smoothing \\
			LLC~\cite{wei2021learn}                 &\xmark &\cmark &\xmark &\xmark & Various& \cmark & \xmark & Differentiable & One-shot & - \\
			FL-AGCNS~\cite{wang2021fl}              &\cmark &\cmark &\xmark &\xmark & Fixed  & \cmark & \xmark & Evolutionary algorithm & One-shot & Federated learning setting\\
            G-Cos~\cite{zhang2021g}                 &\cmark &\xmark &\xmark &\xmark & Fixed  & \cmark & \xmark & Evolutionary algorithm & One-shot & Accelerator search\\
            PAS~\cite{wei2021pooling}               &\cmark &\cmark &\cmark &\xmark & Fixed  & \xmark & \cmark & Differentiable & One-shot & - \\
            FGNAS~\cite{lu2020fgnas}                &\cmark &\xmark &\xmark &\xmark & Fixed  & \cmark & \xmark & RNN controller +RL & - & Software-hardware co-design \\
            GraphPAS~\cite{chen2021graphpas}        &\cmark &\xmark &\xmark &\xmark & Fixed  & \cmark & \xmark & Evolutionary algorithm & Sharing population & Parallel search \\ 
            ALGNN~\cite{cai2021algnn}               &\cmark &\cmark &\xmark &\cmark & Various& \cmark & \xmark & MOPSO~\cite{coello2002mopso} & - &  Consider consumption cost \\ 
            EGNAS~\cite{cai2021edge}                &\cmark &\cmark &\xmark &\xmark & Fixed  & \cmark & \cmark & Differentiable & One-shot & Handle edge features \\
            AutoGEL~\cite{wang2021autogel}          &\cmark &\cmark &\cmark &\xmark & Fixed  & \cmark & \cmark & SNAS~\cite{xie2019snas} & One-shot & Handle edge features \\
            GASSO~\cite{qin2021graph}               &\cmark & \xmark & \xmark & \xmark & Fixed & \cmark & \xmark & Differentiable & One-shot & Graph structure learning\\

G-RNA~\cite{xie2023adversarially} & \cmark & \cmark & \xmark & \xmark & Fixed & \cmark & \xmark & Evolutionary algorithm & One-shot & Search with robustness metrics\\
GRACES~\cite{qin2022graph} & \cmark & \xmark & \cmark & \xmark& Fixed& \xmark & \cmark & Differentiable & One-shot & Handle distribution shifts \\
GAUSS~\cite{guan2022large} & \cmark & \xmark & \xmark & \xmark & Fixed & \cmark & \xmark & Evolutionary algorithm & One-shot & Handle large-scale graphs \\
PasCa~\cite{zhang2022pasca} & \cmark & \xmark & \xmark & \xmark & Various & \cmark & \xmark & Bayesian Optimization & - & Decouple neural message passing \\
PMMM~\cite{li2023differentiable} & \cmark & \cmark & \xmark & \xmark & Fixed & \cmark & \xmark  & Differentiable & One-shot& Search meta paths \\
DHGAS~\cite{zhang2023dynamic} & \cmark & \cmark & \xmark & \xmark& Fixed& \cmark & \xmark & Differentiable & One-shot& Search spatial-temporal modules \\
			\bottomrule
		\end{tabular}
	\label{tab:graphnas}
\end{adjustbox}
\end{table*}
        
NAS methods can be compared in three aspects~\cite{elsken2019neural}: search space, search strategy, and performance estimation strategy. Next, we review NAS methods for graph machine learning from these three aspects and discuss some designs uniquely for graphs. We mainly review NAS for GNNs fitting Eq.~\eqref{eq:mpnn}, which is the focus of the literature. We summarize the characteristics of different methods in Table~\ref{tab:graphnas}. 

\begin{table}
	\centering
  \caption{The typical search space of different types of aggregation weights $a_{ij}$. We omit the layer superscript for brevity. }\label{tab:weights}
	\begin{tabular}{c|l} \toprule
		Type        &  Formulation  \\ \midrule
		CONST       & $a_{ij}^{\text{const}} = 1$  \\
		GCN         & $a_{ij}^{\text{gcn}} =  \frac{1}{\sqrt{\left| \mathcal{N}(i)\right| \left| \mathcal{N}(j) \right| }} $ \\
		GAT         & $a_{ij}^{\text{gat}} = \text{LeakyReLU} \left( \text{ATT} \left(\mathbf{W}_a\left[\mathbf{h}_i, \mathbf{h}_j\right]\right) \right)$ \\
		SYM-GAT     & $a_{ij}^{\text{sym}} = a_{ij}^{\text{gat}}+ a_{ji}^{\text{gat}}$ \\
		COS         & $a_{ij}^{\text{cos}} = \text{cos}\left(\mathbf{W}_a \mathbf{h}_i, \mathbf{W}_a \mathbf{h}_j \right)$ \\
		LINEAR      & $a_{ij}^{\text{lin}} = \text{tanh}\left(\text{sum}\left( \mathbf{W}_a \mathbf{h}_i + \mathbf{W}_a \mathbf{h}_j \right) \right)$ \\
		GENE-LINEAR & $a_{ij}^{\text{gene}} = \text{tanh}\left(\text{sum}\left( \mathbf{W}_a \mathbf{h}_i + \mathbf{W}_a \mathbf{h}_j \right) \right)\mathbf{W}_{a}^\prime $ \\ \bottomrule
	\end{tabular}
\end{table}

\subsection{Search Space}\label{sec:searchspaceold}
The first challenge of NAS on graphs is the search space design since the building blocks of graph machine learning are usually distinct from other deep learning models such as CNNs or RNNs. For GNNs, the search space can be divided into the following five categories.

\spara{Micro search space} 

Following the message-passing framework in Eq.~\eqref{eq:mpnn}, the micro search space defines how nodes exchange messages with others in each layer. Commonly adopted micro search spaces~\cite{ijcai2020-195,zhou2019auto} compose the following components:
\begin{itemize}
	\item Aggregation function $\text{AGG}(\cdot)$: SUM, MEAN, MAX, and MLP.
	\item Aggregation weights $a_{ij}$: common choices are listed in Table~\ref{tab:weights}.
	\item Number of heads when using attentions: 1, 2, 4, 6, 8, 16, etc.
	\item Combining function $\text{COMBINE}(\cdot)$:  CONCAT, ADD, and MLP.
	\item Dimensionality of $\mathbf{h}^{l}$: 8, 16, 32, 64, 128, 256, 512, etc.
	\item Non-linear activation function $\sigma(\cdot)$:  Sigmoid, Tanh, ReLU, Identity, Softplus, Leaky ReLU, ReLU6, and ELU.
\end{itemize}
However, directly searching all these components results in thousands of possible choices in a single message-passing layer. Thus, it may be beneficial to prune the space to focus on a few crucial components depending on applications and domain knowledge~\cite{zhao2020simplifying}.  

\subsubsection{Macro search space} Similar to residual connections and dense connections in CNNs, node representations in one layer of GNNs do not necessarily solely depend on the immediate previous layer~\cite{li2019deepgcns,xu2018representation}. These connectivity patterns between layers form the macro search space. Formally, such designs are formulated as
\begin{equation}
	\mathbf{H}^{(l)} = \sum \nolimits_{j <l} \mathcal{F}_{jl} \left(\mathbf{H}^{(j)}\right),
\end{equation}
where $\mathcal{F}_{jl}(\cdot)$ can be the message-passing layer in Eq.~\eqref{eq:mpnn}, ZERO (i.e., not connecting), IDENTITY (e.g., residual connections), or an MLP. Since the dimensionality of $\mathbf{H}^{(j)}$ can vary, IDENTITY can only be adopted if the dimensionality of each layer matches. 

\subsubsection{Pooling methods} To handle graph-level tasks, information from all the nodes are aggregated to form graph-level representations using the pooling operation in Eq.~\eqref{eq:pool}. Jiang et~al.~\cite{jiang2020graph} propose a pooling search space including row- or column-wise sum, mean, or maximum, attention pooling, attention sum, and flatten. More advanced methods such as hierarchical pooling~\cite{ying2018hierarchical} could also be added to the search space with careful designs. For example,
PAS~\cite{wei2021pooling} further proposes to search for adaptive pooling architectures. Firstly they design a unified framework consisting of four modules: \textit{Aggregation}, \textit{Pooling}, \textit{Read out} and \textit{Merge}, which can cover existing human-designed pooling methods (global and hierarchical) for graph classification. Based on this framework, a novel search space is designed by incorporating popular operations in human-designed architectures. To further enable efficient search, a coarsening strategy is proposed to continuously relax the search space, with the utilization of differentiable search methods. Extensive experiments on six real-world datasets from three domains are conducted, and the results demonstrate the effectiveness and efficiency of the proposed framework.

\subsubsection{Hyper-parameters} Besides architectures, other training hyper-parameters can be incorporated into the search space, i.e., similar to jointly conducting NAS and HPO. Typical hyper-parameters include the learning rate, the number of epochs, the batch size, the optimizer, the dropout rate, and the regularization strengths such as the weight decay. These hyper-parameters can be jointly optimized with architectures or separately optimized after the best-architectures are found. HPO methods in Section~\ref{sec:hpo} can also be combined here.

\subsubsection{Layers} Another critical model choice not incorporated in the above four categories is the number of message-passing layers. Unlike CNNs, most currently successful GNNs are shallow, e.g., with no more than three layers, possibly due to the over-smoothing problem~\cite{li2018deeper,li2019deepgcns}. Limited by this problem, the existing NAS methods for GNNs preset the number of layers as a fixed small number. Except for a recent attempt DeepGNAS~\cite{feng2021search}, how to automatically design deep GNNs while integrating techniques to alleviate over-smoothing remains mostly unexplored.

\subsection{Search Strategy}
The search strategy can be broadly divided into three categories: architecture controllers trained with reinforcement learning (RL), differentiable methods, and evolutionary algorithms. 
\subsubsection{Controller + RL} A widely adopted NAS search strategy is to use a controller to generate the neural architecture descriptions and train the controller with reinforcement learning to maximize the model performance as rewards. For example, if we consider neural architecture descriptions as a sequence, we can use RNNs as the controller~\cite{zoph2017neural}. Such methods can be directly applied to GNNs with a suitable search space and performance evaluation strategy.

\subsubsection{Differentiable} Differentiable NAS methods such as DARTS~\cite{liu2018darts} and SNAS~\cite{xie2019snas} have gained popularity in recent years. Instead of optimizing different operations separately, differentiable methods construct a single super-network (known as the \emph{one-shot model}) containing all possible operations. Formally, we denote
\begin{equation}
	\mathbf{y} = o^{(x,y)}(\mathbf{x}) = \sum \nolimits_{o \in \mathcal{O}} \frac{\exp( \mathbf{z}_{o}^{(x,y)})}{\sum \nolimits_{o^\prime \in \mathcal{O}}\exp( \mathbf{z}_{o^\prime}^{(x,y)})} o(\mathbf{x}), 
\end{equation}
where $o^{(x,y)}(\mathbf{x})$ is an operation in the GNN with input $\mathbf{x}$ and output $\mathbf{y}$, $\mathcal{O}$ are all candidate operations, and $\mathbf{z}^{(x,y)}$ are learnable vectors to control which operation is selected. Briefly speaking, each operation is regarded as a probability distribution of all possible operations.
In this way, the architecture and model weights can be jointly optimized via gradient-based algorithms. The main challenges lie in making the NAS algorithm differentiable, where several techniques such as Gumbel-softmax~\cite{jang2017categorical} and concrete distribution~\cite{maddison2017concrete} are resorted to. When applied to GNNs, slight modification may be needed to incorporate the specific operations defined in the search space, but the general idea of differentiable methods remains unchanged.

\subsubsection{Evolutionary Algorithms} Evolutionary algorithms are a class of optimization algorithms inspired by biological evolution. For NAS, randomly generated architectures are considered initial individuals in a population. Then, new architectures are generated using mutations and crossover operations based on the population. The architectures are evaluated and selected to form the new population, and the same process is repeated. The best architectures are recorded while updating the population, and the final solutions are obtained after sufficient updating steps.   

For GNNs, regularized evolution (RE) NAS~\cite{real2019regularized} has been widely adopted. RE's core idea is an aging mechanism, i.e., in the selection process, the oldest individuals in the population are removed. 
Genetic-GNN~\cite{shi2020bridging} also proposes an evolution process to alternatively update the GNN architecture and the learning hyper-parameters to find the best fit of each other.

\subsubsection{Combinations} It is also feasible to combine these three types of search strategies mentioned above. For example, AGNN~\cite{zhou2019auto} proposes a reinforced conservative search strategy by adopting both RNNs and evolutionary algorithms in the controller and train the controller with RL. By only generating slightly different architectures, the controller can find well-performing GNNs more efficiently.  Peng et~al.~\cite{peng2020learning} adopt CEM-RL~\cite{pourchot2018cemrl}, which combines evolutionary and differentiable methods.

\subsection{Performance Estimation Strategy}
Due to the large number of possible architectures, it is infeasible to fully train each architecture independently. Next, we review some performance estimation strategies. 

A commonly adopted ``trick'' to speed up performance estimation is to reduce fidelity~\cite{elsken2019neural}, e.g., by reducing the number of epochs or the number of data points. This strategy can be directly generalized to GNNs.

Another strategy successfully applied to CNNs is sharing weights among different models, known as parameter sharing or weight sharing~\cite{pham2018efficient}. For differentiable NAS with a large one-shot model, parameter sharing is naturally achieved since the architectures and weights are jointly trained. However, training the one-shot model may be difficult since it contains all possible operations. To further speed up the training process, single-path one-shot model~\cite{guo2020single} has been proposed where only one operation between an input and output pair is activated during each pass.

For NAS without a one-shot model, sharing weights among different architecture is more difficult but not entirely impossible. For example, since it is known that some convolutional filters are common feature extractors, inheriting weights from previous architectures is feasible and reasonable in CNNs~\cite{real2017large}. However, since there is still a lack of understandings of what weights in GNNs represent, we need to be more cautious about inheriting weights~\cite{zhao2020simplifying}. AGNN~\cite{zhou2019auto} proposes three constraints for parameter inheritance: same weight shapes, same attention and activation functions, and no parameter sharing in batch normalization and skip connections.

\subsection{Recent Featured Works}
In this section, we discuss several recent advances in automated graph machine learning that feature in taking topological structure learning, robustness and generalization, scalability, data heterogeneity, efficient architecture search or software-hardware co-design into considerations.

\subsubsection{Architecture Search with Graph Structure Learning} 
Qin~\textit{et al.}~\cite{qin2021graph} investigate the important question that \textit{how NAS is able to select the desired GNN architectures} by conducting a measurement study with experiments, which discovers that gradient based NAS methods tend to select proper architectures based on the usefulness of different types of information with respect to the target task. The explorations further show that gradient based NAS also suffers from noises hidden in the graph, resulting in searching suboptimal GNN architectures. Based on these findings, they propose a Graph differentiable Architecture Search model with Structure Optimization (GASSO), which allows differentiable search of the architecture with gradient descent and is able to discover graph neural architectures with better performance through employing graph structure learning as a denoising process in the search procedure. The proposed GASSO model is capable of simultaneously searching the optimal architecture and adaptively adjusting graph structure by jointly optimizing graph architecture search and graph structure denoising. Extensive experiments on real-world graph datasets demonstrate that the proposed GASSO model is able to achieve state-of-the-art performance compared with existing baselines.

\subsubsection{Robust and Generalizable Graph NAS}
G-RNA~\cite{xie2023adversarially} enhances the robustness of graph NAS methods against adversarial attacks. G-RNA designs a robust search space for the message-passing mechanism by incorporating graph structure mask operations. The graph structure mask operations cover important robust essences of graph structure and could recover various existing defense methods as well. The framework also defines a robustness metric to guide the search process and filter robust architectures. Specifically, G-RNA uses an attack proxy to produce several adversarial samples based on the clean graph, and it searches robust GNNs using the robustness metric with clean and generated adversarial samples.

Besides the robustness, GRACES~\cite{qin2022graph} addresses the limitations of existing graph neural architecture search methods in dealing with distribution shifts between training and test graphs. GRACES tailors a customized GNN architecture suitable for each graph instance to handle the distribution shifts. GRACES uses a self-supervised disentangled graph encoder~\cite{li2021disentangled,li2022disentangled} to characterize invariant factors hidden in diverse graph structures. It further proposes a prototype based architecture self-customization strategy to tailor specialized GNN architectures for graphs based on the similarities of their representations with operation prototypes vectors in the latent space. Finally, the customized supernet provides differentiable weights on the mixture of different operations. GRACES model can be easily optimized in an end-to-end fashion through gradient based methods

\subsubsection{Large-scale Graph NAS}
Guan~\textit{et al.}~\cite{guan2022large} presents the graph architecture search at scale (GAUSS) method, designed to address the limitations of existing graph NAS approaches in handling large-scale graphs. Traditional graph NAS methods are computationally intensive and suffer from the consistency collapse issues, making them unsuitable for large graphs. GAUSS tackles these problems by introducing an efficient light-weight supernet and a joint architecture-graph sampling technique. Specifically, it proposes a graph sampling-based single-path one-shot supernet to minimize computational load. To handle consistency issues, the method incorporates a unique architecture peer learning mechanism on sampled sub-graphs, as well as an architecture importance sampling algorithm. These innovations aim to smooth the highly non-convex optimization objective and stabilize the architecture sampling process. Theoretical analyses and empirical tests on five different datasets, ranging from $10^4$ to $10^8$ vertices, show that GAUSS outperforms existing GNAS methods, marking it as the first framework capable of efficiently handling large-scale graphs with billions of edges within a single GPU day.

Zhang~\textit{et al.}~\cite{zhang2022pasca} further introduces PasCa, a new system aimed at systematically exploring the design space for scalable graph neural networks. It addresses the limitations of current GNNs that are not well-suited for scalability. Through the deconstruction of the neural message-passing mechanism, the authors propose a novel scalable graph neural architecture paradigm (SGAP) that includes a design space with up to 150,000 different architectures. To navigate this expansive design space, it also presents an auto-search engine capable of multi-objective optimization to find GNN architectures that are both efficient and accurate. Empirical studies across ten benchmark datasets reveal that the architectures discovered by PasCa, specifically PasCa-V3, not only offer competitive predictive accuracy but also achieve up to 28.3 times faster training speeds compared to state-of-the-art methods like JK-Net~\cite{xu2018representation}.

\subsubsection{Heterogeneous Graph NAS}
Heterogeneous information networks (HINs) are used to describe real-world data with intricate entities and relationships. Several recent works~\cite{ding2021diffmg,gao2023hgnaspp,li2023differentiable} study the neural architecture search on HINs to handle the heterogeneous node types and relationships. Specifically, Li~\textit{et al.}~\cite{li2023differentiable} proposes a method called Partial Message Meta Multigraph search (PMMM) for automatically optimizing the neural architecture design on Heterogeneous Information Networks (HINs), aiming at better stability and flexibility. It adopts an efficient differentiable framework to search for a meaningful meta multigraph, which captures more flexible and complex semantic relations than a meta graph. The authors also propose a stable algorithm called partial message search to ensure that the searched meta multigraph consistently outperforms manually designed meta-structures. Experimental results on six benchmark datasets for node classification and recommendation tasks demonstrate the effectiveness of PMMM. The proposed method outperforms state-of-the-art heterogeneous GNNs, discovers meaningful meta multigraphs, and exhibits significantly improved stability.

Zhang~\textit{et al.}~\cite{zhang2023dynamic} first consider the temporal information on heterogeneous graphs, and propose a method called Dynamic Heterogeneous Graph Attention Search (DHGAS) for automating the design of Dynamic Heterogeneous Graph Neural Networks (DHGNNs). The existing DHGNNs are manually designed and lack adaptability to diverse dynamic heterogeneous graph scenarios. To overcome this limitation, the authors introduce a search space that considers spatial-temporal dependencies and heterogeneous interactions in graphs and develop an efficient search algorithm. The proposed DHGAS method can automatically discover optimal DHGNN architectures without human guidance. The authors introduce a unified dynamic heterogeneous graph attention (DHGA) framework that enables nodes to attend to their heterogeneous and dynamic neighbors. They also design a localization space to determine attention application and a parameter

\subsubsection{Graph Transformer Architecture Search} 
Besides searching the architecture of graph neural networks, AutoGT~\cite{zhang2023autogt} is proposed to search the architecture of graph transformer, which is another type of strong graph encoder~\cite{ying2021do}. However, unlike their applications in text and image data, using Transformers for graph data involves additional complexities due to the non-euclidean nature of graphs. The authors identify two main challenges. The first challenge is how to creat a unified search space for graph Transformers. The second challenge is how to handle the complex relationship between the architecture of each Transformer layer and graph encoding strategies. To address these challenges, they introduce Automated Graph Transformer (AutoGT), a neural architecture search framework designed for graphs. AutoGT uses a unified graph Transformer formulation and includes a comprehensive search space that considers both architectural and encoding options. To manage the coupling between architecture and graph encodings, the authors propose an encoding-aware performance estimation strategy. Through rigorous experiments, they demonstrate that AutoGT outperforms state-of-the-art hand-crafted models, establishing its effectiveness and broad applicability.

\subsubsection{Search of Efficient Architectures} 
G-Cos~\cite{zhang2021g} is a GNN and accelerator co-search framework that can automatically search for matched GNN structures and accelerators to maximize both task accuracy and acceleration efficiency. Specifically, G-CoS integrates two major components: i) a generic GNN accelerator search space which is applicable to various GNN structures and ii) a one-shot GNN and accelerator co-search algorithm that enables simultaneous and efficient search for optimal GNN structures as well as their matched accelerators. Extensive experiments and ablation studies show that the GNNs together with accelerators generated by G-CoS consistently outperforms state-of-the-art GNNs and GNN accelerators in terms of both task accuracy and hardware efficiency, while only requiring a few hours for the end-to-end generation of the best matched GNNs and their accelerators. Similarly, LPGNAS~\cite{zhao2020learned} jointly searches for architectures and quantisation choices so that both model and buffer sizes can be greatly reduced while keeping similar accuracy as other methods. Their empirical results show that 4-bit weights, 8-bit activations quantisation strategy might be the key for GNNs. ALGNN~\cite{cai2021algnn} considers the computation cost and complexity of the searched model using a multi-objective optimization method.

\subsubsection{Co-design of Software and Hardware}
Lu~\textit{et al.}~\cite{lu2020fgnas} propose FGNAS as the first software-hardware co-design framework for automating the search and deployment of GNNs. Using FPGA as the target platform, the FGNAS framework is able to perform the FPGA-aware graph neural architecture search. FPGA is employed as the vehicle for illustration and implementation of the methods. Specific hardware constraints are considered so that quantization is adopted to compress the model. Under specific hardware constraints, they show the FGNAS framework can successfully identify a solution of higher accuracy while using shorter time than random search and the traditional two-step tuning. To evaluate the design, they conduct experiments on benchmark datasets, i.e., Cora, CiteSeer and PubMed, and the results show that the proposed FGNAS framework has better capability in optimizing the accuracy of GNNs when the hardware implementation is specifically constrained.

\subsection{Discussions}
In this section, we will discuss other unique NAS designs for graphs in terms of search space, transferability and scalability.

\subsubsection{The search space}
Besides the basic search space presented in Section~\ref{sec:searchspaceold}, different graph tasks may require other search spaces. For example,  meta-paths are critical for heterogeneous graphs~\cite{ding2020propagation}, edge features are essential in modeling molecular graphs~\cite{jiang2020graph} and many graph tasks~\cite{wang2021autogel,cai2021edge}, and spatial-temporal modules are needed in skeleton-based recognition~\cite{peng2020learning} and traffic forcasting~\cite{pan2021autostg}. A suitable search space usually requires careful designs and domain knowledge.

\subsubsection{Transferability} It is non-trivial to transfer GNN architectures across different datasets and tasks due to the complexity and diversity of graph tasks. GraphGym~\cite{you2020graph} propose to adopt a fixed set of GNNs as anchors and rank the performance of these GNNs on different tasks and datasets. 
Then, the rank correlation serves as a metric to measure the similarities between different datasets and tasks. The best-performing GNNs of the most similar tasks are transferred to solve the target tasks. 

\subsubsection{The scalability challenge of large-scale graphs} Similar to AutoNE introduced in Section~\ref{sec:hpo}, EGAN~\cite{zhao2021efficient} proposes to sample small graphs as proxies and conduct NAS on the sampled subgraphs to improve the efficiency of NAS. While achieving some progress, more advanced and principle approaches are further needed to handle billion-scale graphs.

\section{Libraries for Automated Graph Machine Learning}
Although there have been quite a few libraries for both graph machine learning and automated machine learning, there is no but one library for automated graph machine learning. 
Therefore, we will briefly overview libraries for graph machine learning and automated machine learning, followed by the in-depth introduction of the world's first dedicated open-source automated graph machine learning library, AutoGL.

\subsection{Libraries for Graph Machine Learning and Automated Machine Learning}\label{sec:tool}

Publicly available libraries are important to facilitate and advance the research and applications of AutoML on graphs. First, we briefly list libraries for graph machine learning and automated machine learning, respectively.

\spara{Graph Machine Learning Libraries}
Popular libraries for graph machine learning include PyTorch Geometric~\cite{Fey2019fast}, Deep Graph Library~\cite{dgl}, GraphNets~\cite{graphnet}, AliGraph~\cite{aligraph}, Euler~\cite{euler}, PBG~\cite{pbg}, PGL~\cite{pgl}, TF-GNN~\cite{tfgnn},  Stellar Graph~\cite{StellarGraph}, Spektral~\cite{spektral}, CodDL~\cite{cogdl}, OpenNE~\cite{openne}, OpenHGNN~\cite{openhgnn}, GEM~\cite{gem}, Karateclub~\cite{karateclub} and classical NetworkX~\cite{hagberg2008exploring}. However, these libraries do not support AutoML.

\spara{Automated Machine Learning Libraries}
On the other hand, AutoML libraries such as NNI~\cite{nni}, AutoKeras~\cite{autokeras}, AutoSklearn~\cite{autosklearn}, Hyperopt~\cite{hyperopt}, TPOT~\cite{tpot}, AutoGluon~\cite{agtabular}, MLBox~\cite{mlbox}, and MLJAR~\cite{mljar} are widely adopted. Unfortunately, because of the uniqueness and complexity of graph tasks, they cannot be directly applied to automate graph machine learning.

Despite their successes, integrating these libraries to fully support automated graph machine learning is non-trivial. This motivates us to design a specific library tailored for automated graph machine learning.

\subsection{AutoGL: A Library for Automated Graph Machine Learning}\label{sec:autogl}

\begin{figure*}[t]
\begin{center}
\includegraphics[width=0.99\textwidth]{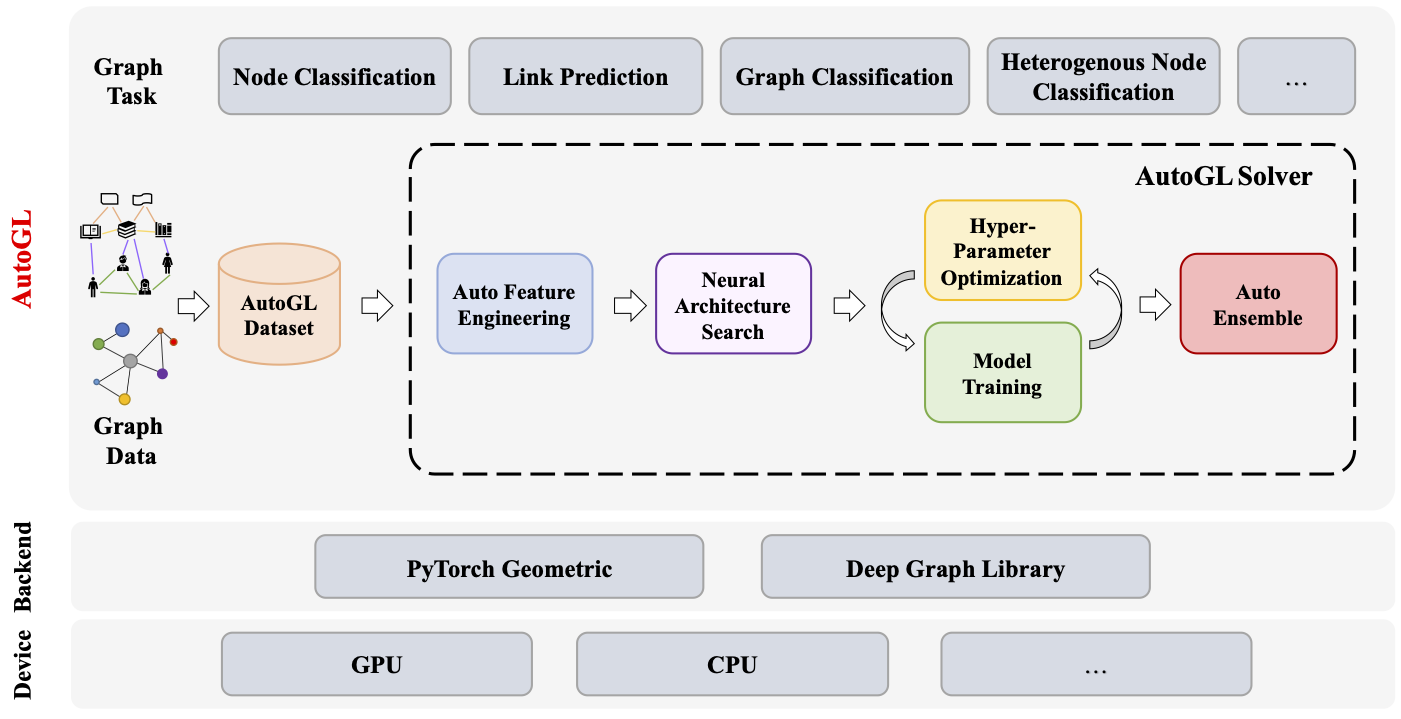}
\end{center}
\caption{The overall framework of AutoGL.}
\label{fig:workflow}
\end{figure*}

To fill this gap, we present Automated Graph Learning (AutoGL)\footnote{This manuscript is based on AutoGL v0.2.0-pre released on 11st, July 2021. Pleases visit the website for the most up-to-the version. }, the first dedicated framework and library for automated graph machine learning. The overall framework of AutoGL is shown in Figure~\ref{fig:workflow}. We summarize and abstract the pipeline of AutoML on graphs into five modules: auto feature engineering, neural architecture search, model training, hyper-parameter optimization, and auto ensemble. For each module, we provide plenty of state-of-the-art algorithms, standardized base classes, and high-level APIs for easy and flexible customization. The AutoGL library is built upon PyTorch Geometric (PyG)~\cite{Fey2019fast}, a widely adopted graph machine learning library. AutoGL has the following key characteristics: 
\begin{itemize}
    \item \textbf{Open source:} The code\footnote{
    \url{https://github.com/THUMNLab/AutoGL/}
    } and detailed documentation\footnote{
    \url{https://autogl.readthedocs.io/}
    } are available online.
    \item \textbf{Easy to use:} AutoGL is designed to be user-friendly. Users can conduct quick AutoGL experiments with less than ten lines of code.
    \item \textbf{Flexible to be extended:} The modular design, high-level base class APIs, and extensive documentation of AutoGL allow flexible and easy customized extensions.
\end{itemize}

\subsection{Detailed Designs of AutoGL}
In this section, we introduce AutoGL designs in detail. AutoGL is designed in a modular and object-oriented fashion to enable clear logic flows, easy usages, and flexible extensions. All the APIs exposed to users are abstracted in a high-level fashion to avoid redundant re-implementation of models, algorithms, and train/evaluation protocols. All the five main modules, i.e., auto feature engineering, neural architecture search, model training, hyper-parameter optimization and auto ensemble, have taken into account the unique characteristics of graph machine learning. Next, we elaborate on the detailed designs for each module.

\subsubsection{AutoGL Dataset} 

We first briefly introduce our dataset management. AutoGL Dataset is currently based on \texttt{Dataset} from PyTorch Geometric and supports common benchmarks for node and graph classification, including the recent Open Graph Benchmark~\cite{hu2020open}. 
We present the complete list of datasets in Table~\ref{tab:dataset}, and
users can also easily customize datasets following our documentation. 

Specifically, we provide widely adopted node classification datasets including Cora, CiteSeer, PubMed~\cite{sen2008collective}, Amazon Computers, Amazon Photo, Coauthor CS, Coauthor Physics~\cite{shchur2018pitfalls}, Reddit~\cite{hamilton2017inductive}, and graph classification datasets such as MUTAG~\cite{mutag}, PROTEINS~\cite{borgwardt2005protein}, IMDB-B, IMDB-M, COLLAB~\cite{graphkernel}, etc. Datasets from Open Graph Benchmark~\cite{hu2020open} are also supported.

\begin{table*}[th]
    \centering
    \caption{The statistics of the supported datasets. For datasets with more than one graph, \#Nodes and \#Edges are the average numbers of all the graphs. \#Features correspond to node features by default, and edge features are specified.}
    \begin{tabular}{lllllll}
    \toprule
    Dataset          & Task                & \#Graphs & \#Nodes & \#Edges & \#Features & \#Classes \\ \midrule
    Cora             & Node  & 1  & 2,708    & 5,429    & 1,433 & 7   \\
    CiteSeer         & Node  & 1  & 3,327    & 4,732    & 3,703 & 6  \\ 
    PubMed           & Node  & 1  & 19,717   & 44,338   & 500   & 3  \\ 
    Reddit           & Node  & 1  & 232,965  &11,606,919& 602   & 41 \\
    Amazon Computers & Node  & 1  & 13,381   & 245,778  & 767   & 10 \\
    Amazon Photo     & Node  & 1  & 7,487    & 119,043  & 745   & 8  \\
    Coauthor CS      & Node  & 1  & 18,333   & 81,894   & 6,805 & 15 \\
    Coauthor Physics & Node  & 1  & 34,493   & 247,962  & 8,415 & 5  \\
    ogbn-products    & Node  & 1  & 2,449,029  & 61,859,140    & 100        & 47 \\
    ogbn-proteins    & Node  & 1  & 132,534    & 39,561,252    & 8(edge) & 112\\
    ogbn-arxiv       & Node  & 1  & 169,343    & 1,166,243     & 128     & 40 \\
    ogbn-papers100M  & Node  & 1  &111,059,956 & 1,615,685,872 & 128     & 172\\
    Mutag            & Graph & 188     & 17.9  & 19.8    & 7s & 2 \\
    PTC              & Graph & 344     & 14.3  & 14.7    & 18 & 2 \\
    ENZYMES          & Graph & 600     & 32.6  & 62.1    & 3 & 6 \\
    PROTEINS         & Graph & 1,113   & 39.1  & 72.8    & 3 & 2 \\
    NCI1             & Graph & 4,110   & 29.8  & 32.3    & 37 & 2 \\
    COLLAB           & Graph & 5,000   & 74.5  & 2,457.8 & - & 3 \\
    IMDB-B           & Graph & 1,000   & 19.8  & 96.5    & - & 2 \\
    IMDB-M           & Graph & 1,500   & 13.0  & 65.9    & - & 3 \\ 
    REDDIT-B    & Graph & 2,000   & 429.6 & 497.8   & - & 2 \\ 
    REDDIT-MULTI5K   & Graph & 5,000   & 508.5 & 594.9   & - & 5 \\
    REDDIT-MULTI12K  & Graph & 11,929  & 391.4 & 456.9   & - & 11\\
    ogbg-molhiv      & Graph & 41,127  & 25.5  & 27.5    & 9, 3(edge) & 2 \\
    ogbg-molpcba     & Graph & 437,929 & 26.0  & 28.1    & 9, 3(edge) & 128\\
    ogbg-ppa         & Graph & 158,100 & 243.4 & 2,266.1 & 7(edge) & 37 \\
    \bottomrule
    \end{tabular}
    \label{tab:dataset}
\end{table*}

\subsubsection{Auto Feature Engineering} 

The graph data is first processed by the auto feature engineering module, where various nodes, edges, and graph-level features can be automatically added, compressed, or deleted to help boost the graph learning process after. Graph topological features can also be extracted to utilize graph structures better.

Currently, we support 24 feature engineering operations abstracted into three categories: generators, selectors, and graph features. The generators aim to create new node and edge features based on the current node features and graph structures. The selectors automatically filter out and compress features to ensure they are compact and informative. Graph features focus on generating graph-level features. 

We summarize the supported generators in Table~\ref{tab:generators}, including Graphlets~\cite{milo2002network}, EigenGNN~\cite{zhang2021eigen}, PageRank~\cite{pagerank}, local degree profile, normalization, one-hot degrees, and one-hot node IDs. For selectors, GBDT~\cite{NIPS2017_6449f44a} and FilterConstant are supported. An automated feature engineering method DeepGL~\cite{rossi2018deep} is also supported, functioning as both a generator and a selector. For graph feature, Netlsd~\cite{tsitsulin2018netlsd} and a set of graph feature extractors implemented in NetworkX~\cite{hagberg2008exploring} are wrapped, e.g.,
\texttt{NxTransitivity},  \texttt{NxAverageClustering}, etc.

We also provide convenient wrappers that support feature engineering operations in PyTorch Geometric~\cite{Fey2019fast} and NetworkX~\cite{hagberg2008exploring}. 
Users can easily customize feature engineering methods by inheriting from the class \texttt{BaseGenerator}, \texttt{BaseSelector}, and \texttt{BaseGraph}, or \texttt{BaseFeatureEngineer} if the methods do not fit in our categorization.

\begin{table}[h]
\caption{Supported generators in the auto feature engineering module.}
\label{tab:generators}
\centering
\begin{tabular}{l|l}
\toprule
    Name & Description \\ \midrule
    \texttt{graphlet} & Local graphlet numbers\\ 
    \texttt{eigen}    & EigenGNN features. \\ 
    \texttt{pagerank} & PageRank scores. \\ 
    \texttt{PYGLocalDegreeProfile} & Local Degree Profile features \\ 
    \texttt{PYGNormalizeFeatures} & Row-normalize all node features \\
    \texttt{PYGOneHotDegree} & One-hot encoding of node degrees. \\ 
    \texttt{onehot} & One-hot encoding of node IDs \\ \bottomrule
\end{tabular}
\end{table}

\subsubsection{Neural Architecture Search} 

In AutoGL, Neural Architecture Search (NAS) aims to automate the construction of Graph Neural Networks. The best GNN model will be searched using various NAS methods to fit the current datasets. In Neural Architecture Search module, Algorithm, GNNSpace, and Estimator submodule is developed to further solve the search problem. GNNSpace defines the whole search range where we explore the best models. Algorithms are used to determine which architectures should be evaluated next, and the Estimators are used for deriving the performances of target architectures.

\textcolor{black}{We have supported various NAS models, including algorithms specified for graph data like AutoNE~\cite{tu2019autone} and AutoGR~\cite{wang2021explainable} and general-purpose algorithms like random search~\cite{bergstra2012random}, Tree Parzen Estimator~\cite{bergstra2011algorithms}, etc. Users can customize HPO algorithms by inheriting from the \texttt{BaseHPOptimizer} class.}

\subsubsection{Model Training} 

This module handles the training and evaluation process of graph machine learning tasks with two functional sub-modules: Model and Trainer. Model handles the construction of graph machine learning models, e.g., GNNs, by defining learnable parameters and the forward pass. Trainer controls the optimization process for the given model. Common optimization methods are packaged as high-level APIs to provide neat and clean interfaces. More advanced training controls and regularization methods in graph tasks like early stopping and weight decay are also supported.

The model training module supports both node-level and graph-level tasks, e.g., node classification and graph classification. Commonly used models for node classification such as GCN~\cite{kipf2017semi}, GAT~\cite{velickovic2018graph}, and GraphSAGE~\cite{hamilton2017inductive}, GIN~\cite{xu2019powerful}, and pooling methods such as Top-K Pooling~\cite{topk} are supported. Users can quickly implement their own graph models by inheriting from the \texttt{BaseModel} class and add customized tasks or optimization methods by inheriting from \texttt{BaseTrainer}.

\subsubsection{Hyper-Parameter Optimization} 

The Hyper-Parameter Optimization (HPO) module aims to automatically search for the best hyper-parameters of a specified model and training process, including but not limited to architecture hyper-parameters such as the number of layers, the dimensionality of node representations, the dropout rate, the activation function, and training hyper-parameters such as the learning rate, the weight decay, the number of epochs. The hyper-parameters, their types (e.g., integer, numerical, or categorical), and feasible ranges can be easily set.

We have supported various HPO algorithms, including algorithms specified for graph data like AutoNE~\cite{tu2019autone} and AutoGR~\cite{wang2021explainable} and general-purpose algorithms like random search~\cite{bergstra2012random}, Tree Parzen Estimator~\cite{bergstra2011algorithms}, etc. Users can customize HPO algorithms by inheriting from the \texttt{BaseHPOptimizer} class.

\subsubsection{Auto Ensemble} 

This module can automatically integrate the optimized individual models to form a more powerful final model. Currently, we have adopted two kinds of ensemble methods: voting and stacking. Voting is a simple yet powerful ensemble method that directly averages the output of individual models. Stacking trains another meta-model to combine the output of models. We have supported general linear models (GLM) and gradient boosting machines (GBM) as meta-models. 

\subsubsection{AutoGL Solver} 
On top of the modules mentioned above, we provide another high-level API Solver to control the overall pipeline. In Solver, the five modules are integrated systematically to form the final model. Solver receives the feature engineering module, a model list, the HPO module, and the ensemble module as initialization arguments to build an Auto Graph Learning pipeline. Given a dataset and a task, Solver first perform auto feature engineering to clean and augment the input data, then optimize all the given models using the model training and HPO module. At last, the optimized best models will be combined by the Auto Ensemble module to form the final model.

Solver also provides global controls of the AutoGL pipeline. For example, the time budget can be explicitly set to restrict the maximum time cost, and the training/evaluation protocols can be selected from plain dataset splits or cross-validation.

\subsection{Evaluation of AutoGL}

In this section, we provide experimental results. Note that we mainly want to showcase the usage of AutoGL and its main functional modules rather than aiming to achieve the new state-of-the-art on benchmarks or compare different algorithms. 
For node classification, we use Cora, CiteSeer, and PubMed with the standard dataset splits from~\cite{kipf2017semi}. 
For graph classification, we follow the setting in~\cite{errica2020fair} and report the average accuracy of 10-fold cross-validation on MUTAG, PROTEINS, and IMDB-B.

\spara{AutoGL Results} 

We turn on all the functional modules in AutoGL, and report the fully automated results in Table \ref{semi} and Table \ref{graph}. We use the best single model for graph classification under the cross-validation setting. We observe that in all the benchmark datasets, AutoGL achieves better results than vanilla models, demonstrating the importance of AutoML on graphs and the effectiveness of the proposed pipeline in the released library.

\begin{table}[ht]
\centering
\caption{The results of node classification}
\begin{tabular}{cccc}
\toprule
Model & Cora & CiteSeer & PubMed \\
\midrule
GCN & $80.9 \pm 0.7$ & $70.9 \pm 0.7$ & $78.7 \pm 0.6$ \\
GAT & $82.3 \pm 0.7$ & $71.9 \pm 0.6$ & $77.9 \pm 0.4$ \\
GraphSAGE & $74.5 \pm 1.8$ & $67.2 \pm 0.9$ & $76.8 \pm 0.6$ \\
AutoGL & $\mathbf{83.2 \pm 0.6}$ & $\mathbf{72.4 \pm 0.6}$ & $\mathbf{79.3 \pm 0.4}$ \\
\bottomrule
\label{semi}
\end{tabular}
\end{table}

\begin{table}[ht]
\centering
\caption{The results of graph classification}
\begin{tabular}{cccc}
\toprule
Model & MUTAG & PROTEINS & IMDB-B \\
\midrule
Top-K Pooling & $80.8 \pm 7.1$ & $69.5 \pm 4.4$ & $71.0 \pm 5.5$ \\
GIN & $82.7 \pm 6.9 $ & $66.5 \pm 3.9$ & $69.1 \pm 3.7$ \\
AutoGL & $\mathbf{87.6 \pm 6.0}$ & $\mathbf{73.3 \pm 4.4}$ & $\mathbf{72.1 \pm 5.0}$ \\
\bottomrule
\label{graph}
\end{tabular}
\end{table}

\spara{Hyper-Parameter Optimization} 

\begin{table*}[t]
\centering
\caption{The results of different HPO methods for node classification}
\begin{tabular}{cccccccc}
\toprule
\multicolumn{2}{c}{} & \multicolumn{2}{c}{Cora} & \multicolumn{2}{c}{CiteSeer} & \multicolumn{2}{c}{PubMed} \\
Method & Trials & GCN & GAT & GCN & GAT & GCN & GAT \\
\midrule
\multicolumn{2}{c}{None} & $80.9\pm 0.7$ & $82.3\pm0.7$ & $70.9\pm0.7$ & $71.9\pm 0.6$ & $78.7\pm0.6$ & $77.9\pm0.4$ \\
\midrule
\multirow{3}{*}{random} & 1 & $81.0 \pm 0.6$ & $81.4 \pm 1.1$ & $70.4 \pm 0.7$ & $70.1 \pm 1.1$ & $78.3 \pm 0.8$ & $76.9 \pm 0.8$ \\
& 10 & $82.0\pm0.6$ & $82.5\pm0.7$ & $71.5\pm0.6$ & $\bf 72.2\pm0.7$ & $79.1\pm0.3$ & $78.2\pm0.3$ \\
& 50 & $81.8\pm1.1$ & $\bf 83.2\pm0.7$ & $71.1\pm1.0$ & $72.1\pm1.0$ & $\bf 79.2\pm0.4$ & $78.2\pm0.4$ \\
\midrule
\multirow{3}{*}{TPE} & 1 & $81.8 \pm 0.6$ & $81.9 \pm 1.0$ & $70.1 \pm 1.2$ & $71.0 \pm 1.2$ & $78.7 \pm 0.6$ & $77.7 \pm 0.6$ \\
& 10 & $82.0 \pm 0.7$ & $82.3 \pm 1.2$ & $71.2 \pm 0.6$ & $72.1 \pm 0.7$ & $79.0 \pm 0.4$ & $\bf 78.3 \pm 0.4$ \\
& 50 & $\bf 82.1 \pm 1.0$ & $83.2 \pm 0.8$ & $\bf 72.4 \pm 0.6$ & $71.6 \pm 0.8$ & $79.1 \pm 0.6$ & $78.1 \pm 0.4$ \\
\bottomrule
\label{hpo:node}
\end{tabular}
\end{table*}

\begin{table*}
\centering
\caption{The results of different HPO methods for graph classification}
\begin{tabular}{ccccccc}
\toprule
\multirow{2}{*}{HPO} & \multicolumn{2}{c}{MUTAG} & \multicolumn{2}{c}{PROTEINS} & \multicolumn{2}{c}{IMDB-B} \\
 & Top-K Pooling & GIN & Top-K Pooling & GIN & Top-K Pooling & GIN \\
\midrule
None & $76.3\pm7.5$ & $82.7\pm6.9$ & $69.5\pm4.4$ & $66.5\pm3.9$ & $71.0\pm5.5$ & $69.1\pm3.7$\\
random & $82.7\pm 6.8$ & $\bf 87.6 \pm 6.0$ & $\bf 73.3\pm4.4$ & $\bf 71.0\pm5.9$ & $71.5\pm 4.1$ & $\bf 71.3\pm 4.0$ \\
TPE & $\mathbf{ 83.9\pm 10.1}$ & $86.7 \pm 6.2$ & $72.3\pm5.5$ & $71.0\pm7.2$ & $\bf 71.6\pm 2.5$ & $70.2\pm 3.7$ \\
\bottomrule
\label{hpo:graph}
\end{tabular}

\end{table*}

Table~\ref{hpo:node} reports the results of two implemented HPO methods, i.e., random search and TPE~\cite{bergstra2011algorithms}, for the semi-supervised node classification task. As shown in the table, as the number of trials increases, both HPO methods tend to achieve better results. Besides, both methods outperform vanilla models without HPO. Note that a larger number of trials do not guarantee better results because of the potential overfitting problem. We further test these HPO methods with ten trials for the graph classification task and report the results in Table \ref{hpo:graph}. The results generally show improvements over the default hand-picked parameters on all datasets.

\spara{Auto Ensemble} 

\begin{table}[h]
\centering
\caption{The performance of the ensemble module of AutoGL for the node classification task.} 
\label{tab:results_ensemble}
  \centering
  \begin{tabular}{lcccccc}
  \toprule
  Base Model & Cora & CiteSeer & PubMed \\ 
  \midrule
  GCN & $81.1 \pm 0.9$ & $69.6\pm 1.1$ & $\bf 78.5 \pm 0.4 $ \\
  GAT & $82.0 \pm 0.5$ & $70.4\pm 0.6$ & $77.7 \pm 0.5 $\\
  \midrule
  Ensemble  & $\bf 82.2 \pm 0.4$ & $\bf 70.8 \pm 0.5$ & $\bf 78.5 \pm 0.4$ \\
  \bottomrule
  \end{tabular}
\end{table}

Table~\ref{tab:results_ensemble} reports the performance of the ensemble module as well as its base learners for the node classification task. We use voting as the example ensemble method and choose GCN and GAT as the base learners. The table shows that the ensemble module achieves better performance than both the base learners, demonstrating the effectiveness of the implemented ensemble module.

\subsection{Advanced Functions of AutoGL}
We have presented AutoGL, the first library for automated graph machine learning, which is open-source, easy to use, and flexible to be extended.  
Currently, we are actively developing AutoGL and have supported the following advanced functionalities:
\begin{itemize}
    \item Support more graph models. We have supported self-supervised graph models and robust graph models now.
    \item Handle more graph tasks. We have supported heterogeneous node classification tasks now and plan to support more complex spatial-temporal graphs. 
    \item Support more graph library backends. We have supported Deep Graph Library (DGL)~\cite{dgl} backend including homogeneous node classification, link prediction, and graph classification tasks. AutoGL is also compatible with PyG 2.0~\cite{Fey2019fast} now.
\end{itemize}
All kinds of inputs and suggestions are also warmly welcomed.

\section{Benchmark}\label{sec:benchmark}

To promote the further development of automated graph machine learning, NAS-Bench-Graph~\cite{qin2022bench} is proposed as one benchmark tailored to enable unified, reproducible, and efficient evaluations for graph NAS. Two key issues are identified: i) the lack of a standard experimental setting, making comparisons across research unreliable and non-reproducible, and ii) the computational inefficiency of graph NAS methods. To resolve these issues, NAS-Bench-Graph constructs a unified search space, encompassing 26,206 unique GNN architectures, and offers a standardized evaluation protocol. 
All architectures have been trained and evaluated on nine representative graph datasets, with metrics such as training, validation, and test performance, latency, and the number of parameters recorded. This results in a look-up table that allows for fair and efficient comparisons without additional computational costs. The authors also showcase the benchmark's compatibility with existing GraphNAS libraries like AutoGL and NNI. The work claims to be the first of its kind to offer a benchmark in the domain of GraphNAS.

\subsection{Benchmark Design}

Here we elucidates the methodology employed in the development of our benchmark for graph NAS, encapsulating search space design (Section \ref{sec:searchspace}), utilized datasets (Section \ref{sec:dataset}), and experimental configurations (Section \ref{sec:expset}).

\subsubsection{Search Space Design}\label{sec:searchspace}

\begin{figure*}[h]
  \centering
  \includegraphics[width=0.8\linewidth]{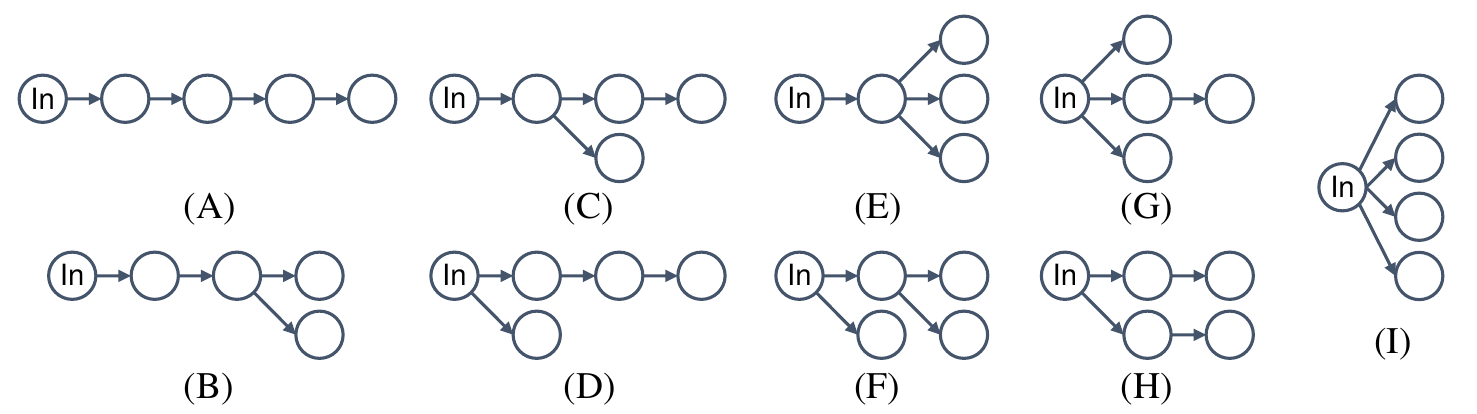}
  \caption{Nine different choices of our macro search space. Each node is a representation of vertices and each edge is an operation~\cite{qin2022bench}.}
  \label{fig:macrospace}
\end{figure*}

To achieve a judicious equilibrium between effectiveness and efficiency, we propose an intricately crafted, yet computationally tractable search space. In particular, we conceptualize the macro search space governing Graph Neural Network (GNN) architectures as a Directed Acyclic Graph (DAG), thereby providing a formal structure for the computational paradigm\footnote{For disambiguation, we refer to nodes and edges in the computational graph, as opposed to vertices and links that constitute the graph data.}. In this DAG, each computational node signifies a vertex representation, while each edge symbolizes a specific operation. The DAG is composed of six nodes, including the input and output nodes. Notably, each intermediate node is restricted to having a single incoming edge. As a result, the aforementioned DAG encompasses nine selectable configurations, as delineated in Figure~\ref{fig:macrospace}. Any intermediate nodes devoid of successor nodes are concatenated to the output node.

In addition to this overarching macro space, we also incorporate optional fully-connected layers in the pre-processing and post-processing layers, following the design principles established in GraphGym~\cite{you2020design} and PasCa~\cite{zhang2022pasca}. To circumvent an exponential increase in the complexity of the search space, the number of layers in both pre-processing and post-processing phases are treated as hyperparameters, a point that will be elaborated upon in Section~\ref{sec:expset}. To culminate the architecture, a task-specific fully-connected layer is employed to generate the model's prediction.

For the set of admissible operations, we focus on seven GNN layers that have garnered widespread acceptance in the literature: GCN~\cite{kipf2017semi}, GAT~\cite{velickovic2018graph}, GraphSAGE~\cite{hamilton2017inductive}, GIN~\cite{xu2019powerful}, ChebNet~\cite{defferrard2016convolutional}, ARMA~\cite{bianchi2021graph}, and k-GNN~\cite{morris2019weisfeiler}. Furthermore, we introduce the Identity operation to facilitate residual connections and a fully-connected layer that does not exploit graph structures.

In summation, the search space we designed is remarkably expansive, consisting of 26,206 unique architectures after accounting for isomorphic structures (i.e., structures that may exhibit different topological characteristics yet perform identical functionalities, further elaborated in Appendix A.5). Importantly, this search space encapsulates a myriad of GNN variants, including but not limited to the previously mentioned methods, and extends to more advanced architectures such as JK-Net~\cite{xu2018representation} and residual- and dense-like GNNs~\cite{li2019deepgcns}.

\begin{table*}[t]
\caption{The hyper-parameters and hardware used for each dataset. \#Pre and \#Post denotes the number of pre-process and post-process layers, respectively.}
\label{tab:hp}
\centering
\begin{adjustbox}{max width=1.0\textwidth}
\begin{tabular}{lrrrlcllr}
\toprule
Dataset          & \#Pre & \#Post & Dimension & Dropout & Optimizer & LR & WD & \# Epoch \\ \midrule
Cora             & 0    & 1      & 256  & 0.7   & SGD & 0.1 & 0.0005 & 400 \\
CiteSeer         & 0    & 1      & 256   & 0.7   & SGD & 0.2 & 0.0005 & 400 \\ 
PubMed           & 0   & 0     & 128     & 0.3   & SGD & 0.2 & 0.0005 & 500 \\ 
Coauthor-CS      & 1   & 0     & 128   & 0.6  & SGD & 0.5 & 0.0005 & 400  \\
Coauthor-Physics & 1 & 1    & 256   & 0.4   & SGD & 0.01 & 0 & 200   \\
Amazon-Photo     & 1    & 0    & 128     & 0.7   & Adam & 0.0002 & 0.0005 & 500  \\
Amazon-Computers & 1   & 1    & 64     & 0.1  & Adam & 0.005 & 0.0005 & 500 \\
ogbn-arxiv       & 0  & 1  & 128 & 0.2    & Adam  & 0.002 & 0 & 500  \\
ogbn-proteins    & 1  & 1 & 256 & 0 & Adam & 0.01 & 0.0005 & 500\\
\bottomrule
\end{tabular}
\end{adjustbox}
\end{table*}

\begin{table*}[t]
\caption{The average training time of  architectures on each dataset.}
\label{tab:time}
\centering
    \begin{tabular}{lr|lr|lr}
    \toprule
    Dataset          & Time & Dataset & Time & Dataset & Time\\ \midrule
    Cora             & 5.8s & Coauthor-CS & 8.6s  & Amazon-Computers &9.8s \\
    CiteSeer         & 6.2s & Coauthor-Physics & 15.4s   & ogbn-arXiv & 71s\\ 
    PubMed           & 7.8s & Amazon-Photo & 8.8s  & ogbn-proteins & 50min \\ 
    \bottomrule
    \end{tabular}
\end{table*}

\subsubsection{Datasets}\label{sec:dataset}
In the course of this study, we employ nine diverse, publicly available datasets extensively utilized in the realm of Graph Neural Architecture Search (GraphNAS). Specifically, these datasets include Cora, CiteSeer, and PubMed~\cite{sen2008collective}, alongside Coauthor-CS, Coauthor-Physics, Amazon-Photo, and Amazon-Computer~\cite{pitfall}, as well as ogbn-arXiv and ogbn-proteins~\cite{ogb}. These datasets span an array of sizes, ranging from several thousands to millions of edges, and encompass diverse application domains such as citation networks, e-commerce graphs, and protein interaction networks. 

Regarding dataset partitioning strategies, we adhere to the publicly available semi-supervised settings for Cora, CiteSeer, and PubMed as delineated by~\cite{yang2016revisiting}. This involves utilizing 20 labeled nodes per class for training and 500 nodes for validation purposes. For the Amazon and Coauthor datasets, we employ a random partitioning scheme for train/validation/test splits in a semi-supervised fashion, as recommended by~\cite{pitfall}. Specifically, each class is represented by 20 nodes for training, 30 nodes for validation, and the remaining nodes are used for testing. In the case of ogbn-arXiv and ogbn-proteins, we abide by the official dataset splits.

Pertaining to the ogbn-proteins dataset, preliminary experiments indicated that the utilization of Graph Isomorphism Network (GIN) and k-Graph Neural Network (k-GNN) operations resulted in parameter explosion, thereby yielding uninterpretable outcomes. Additionally, employing Graph Attention Networks (GAT) and Chebyshev Spectral Graph Convolution Networks (ChebNet) led to out-of-memory errors on our high-capacity GPUs equipped with 32GB of memory. To mitigate these computational inefficiencies, we circumscribed the candidate operations for ogbn-proteins to Graph Convolution Networks (GCN), Attention-based Recurrent Multi-layer Average networks (ARMA), GraphSAGE, Identity mappings, and fully connected layers. Consequently, this constraint yielded a feasible architecture space comprising 2,021 candidate models for ogbn-proteins.

\subsubsection{Settings}\label{sec:expset}
\paragraph{Hyper-parameters} For fair and reproducible comparisons, we propose a unified evaluation protocol and consider the following hyper-parameters with tailored ranges:
\begin{itemize}
    \item Number of pre-process layers: 0 or 1. 
    \item Number of post-process layers: 0 or 1. 
    \item Dimensionality of hidden units: 64, 128, or 256.
    \item Dropout rate: 0.0, 0.1, 0.2, 0.3, 0.4, 0.5, 0.6, 0.7, 0.8.
    \item Optimizer: SGD or Adam.
    \item Learning Rate (LR): 0.5, 0.2, 0.1, 0.05, 0.02, 0.01, 0.005, 0.002, 0.001.
    \item Weight Decay: 0 or 0.0005.
    \item Number of training epochs: 200, 300, 400, 500.
\end{itemize}

For each dataset, we establish fixed hyper-parameters across all architectures in order to ensure a fair comparison. It should be noted that exhaustively enumerating combinations of architectures and hyper-parameters would lead to an impractical number of architecture hyper-parameter pairs in the order of billions. Hence, we adopt a two-step approach, first optimizing the hyper-parameters to a suitable value that can accommodate various Graph Neural Network (GNN) architectures, and subsequently focusing on the GNN architectures themselves. Specifically, we select 30 GNN architectures from our search space as "anchors" and employ random search for hyper-parameter optimization~\cite{bergstra2012random}. The set of 30 anchor architectures comprises 20 randomly chosen architectures from our search space, along with 10 classic GNN architectures including Graph Convolutional Networks (GCN), Graph Attention Networks (GAT), Graph Isomorphism Networks (GIN), GraphSAGE, and ARMA with 2 and 3 layers. We optimize the hyper-parameters by maximizing the average performance of these anchor architectures. The specific hyper-parameters selected for each dataset are shown in Table~\ref{tab:hp}.

\paragraph{Metrics} During the training of each architecture, we record a comprehensive set of metrics covering both model effectiveness and efficiency. These metrics include the loss values and evaluation metric at each epoch for both the training, validation, and testing sets, as well as the model latency and the number of parameters. The hardware and software configurations utilized in our experiments are provided as follows:
\begin{itemize}
  \item Operating System: Ubuntu 18.04.6 LTS for PubMed, ogbn-arXiv, and CentOS Linux release 7.6.1810 for the others.
  \item CPU: Intel(R) Xeon(R) Gold 6129 CPU @ 2.30GHz for PubMed, ogbn-arXiv, and Intel(R) Xeon(R) Gold 6240 CPU @ 2.60GHz for the others.
  \item GPU: NVIDIA GeForce RTX 3090 with 24GB of memories for PubMed, ogbn-arXiv, and NVIDIA Tesla V100 with 16GB of memories for the others.
  \item Software: Python 3.9.12, PyTorch 1.11.0+cu113, PyTorch-Geometric 2.0.4~\cite{Fey2019fast}.
\end{itemize}
Moreover, to account for the inherent variability in the training process, all experiments are repeated three times using different random seeds.
The average training time of architectures on each dataset is reported in Table~\ref{tab:time}.  The total computational cost incurred in creating our benchmark dataset amounts to approximately 8,000 GPU hours.

\subsection{Example Usages}

This section demonstrates the utilization of NAS-Bench-Graph in conjunction with established open-source libraries such as AutoGL~\cite{guan2021autoattend} and NNI~\cite{NNIB}. Specifically, we employ two NAS algorithms, namely GNAS~\cite{gao2020graph}  and Auto-GNN~\cite{zhou2019auto}, within the AutoGL framework. Additionally, we employ Random Search~\cite{li2020random}, Evolutionary Algorithm (EA), and Policy-based Reinforcement Learning (RL) algorithms within NNI. To ensure equitable comparisons, we restrict each algorithm's access to the performance data of only 2\% of the total architectures within the search space. The obtained results are presented in Table \ref{tab:use}. Furthermore, we include the performance metrics of the top 5\% architectures, representing the 20-quantiles for each dataset, within the aforementioned table.

\begin{table*}[!h]
  \caption{The performance of NAS methods in AutoGL and NNI using  NAS-Bench-Graph. The best performance for each dataset is marked in bold. We also show the performance of the top 5\% architecture (i.e., 20-quantiles) as a reference line. The results are averaged over five experiments with different random seeds and the standard errors are shown in the bottom right.}  
  \label{tab:use}
\centering
  \begin{tabular}{l|l|p{0.83cm}p{0.83cm}p{0.83cm}p{0.83cm}p{0.83cm}p{0.83cm}p{0.83cm}p{0.83cm}p{0.83cm}}
  \toprule
  Library                                      & Method & Cora  & CiteSeer & PubMed & CS    & Physics & Photo & Computers & arXiv & proteins \\ \midrule
  \multicolumn{1}{l|}{\multirow{2}{*}{AutoGL}} & GNAS   & 82.04$_{0.17}$  & $\textbf{70.89}_{ 0.16}$  &  77.79$_{ 0.02}$ & 90.97$_{0.06}$  & 92.43$_{0.04}$ & 92.43$_{0.03}$  & 84.74$_{0.20}$  & 72.00$_{0.02}$  &   \textbf{78.71}$_{0.11}$   \\
  \multicolumn{1}{l|}{}                        & Auto-GNN  & 81.80$_{0.00}$  & 70.76$_{0.12}$ & 77.69$_{ 0.16}$ &     $\textbf{91.04}_{0.04}$  &  92.42$_{0.16}$ &  92.38$_{0.01}$ & 84.53$_{0.14}$ &  \textbf{72.13}$_{0.03}$  &  78.54$_{0.30}$   \\ \midrule
  \multirow{3}{*}{NNI}                         & Random & 82.09$_{ 0.08}$ & 70.49$_{0.08}$    & 77.91$_{ 0.07}$  & 90.93$_{ 0.07}$ & 92.35$_{0.05}$   & $\textbf{92.44}_{0.02}$ & 84.78$_{0.14}$     & 72.04$_{0.05}$ & 78.32$_{0.14}$    \\
  & EA     & 81.85$_{ 0.20}$ & 70.48$_{ 0.12}$    & $\textbf{77.96}_{ 0.12}$  & 90.60$_{ 0.07}$ & 92.22$_{ 0.08}$   & 92.43$_{0.02}$ & 84.29$_{0.29}$     & 71.91$_{0.06}$ & 77.93$_{0.21}$   \\
  & RL     & $\textbf{82.27}_{  0.21}$ & 70.66$_{ 0.12}$    & $\textbf{77.96}_{ 0.09}$  & 90.98$_{ 0.01}$ & $\textbf{92.48
  }_{0.03}$   & 92.42$_{0.06}$ & $\textbf{84.90}_{0.19}$     & \textbf{72.13}$_{0.05}$ & 78.52$_{0.18}$    \\ \midrule
      \multicolumn{2}{c|}{The top 5\%} &  80.63 & 69.07 & 76.60 & 90.01 & 91.67 & 91.57 & 82.77 & 71.69 & 78.37     \\
   \bottomrule
  \end{tabular}
  \end{table*}

The detailed example codes are provided as follows. All the codes and recorded metrics for the trained models are available at \url{https://github.com/THUMNLab/NAS-Bench-Graph}. Next, we provide some example usages.

\begin{figure}[h]
    \centering
    \includegraphics[width=\linewidth]{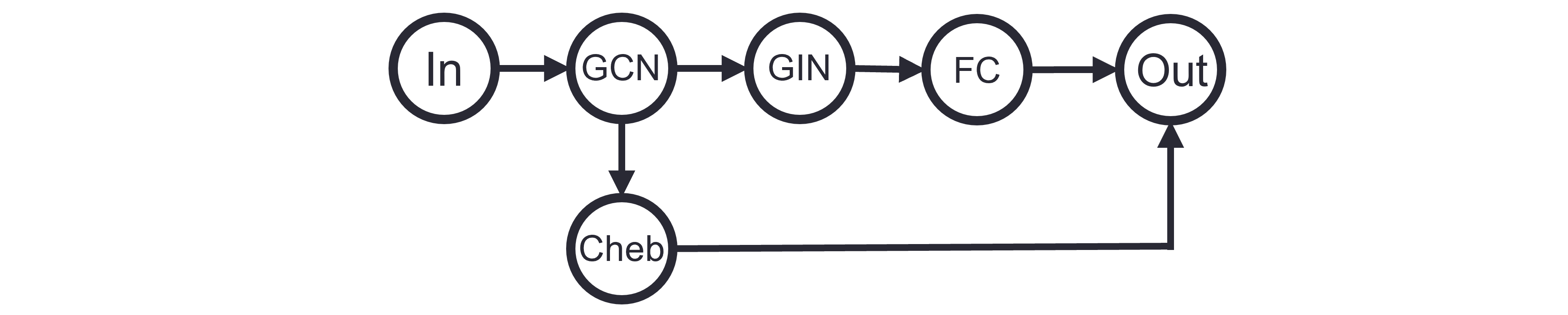}
    \caption{An example architecture.}
    \label{fig:example}
\end{figure}

At first, the benchmark of a certain dataset, e.g., Cora, can be read as:
\begin{lstlisting}
from readbench import lightread
bench = lightread('cora')
\end{lstlisting}
The data is stored as a Python dictionary.
In order to capture the metrics of interest, it is imperative to define an architecture by specifying its macro space and corresponding operations. In our approach, we impose a restriction on the directed acyclic graph (DAG) of the computation graph, dictating that each intermediate node may have only one input node. This constraint allows us to represent the macro space using a list of integers, wherein each integer denotes the index of the input node for a given computing node. Specifically, the value of 0 corresponds to the raw input, 1 corresponds to the first computing node, and so forth. Additionally, the operations associated with the architecture can be described using a list of strings of equal length, providing a concise representation of the architectural choices made.
For example, to specify the architecture shown in Figure \ref{fig:example}, we can use the following code:
\begin{footnotesize}
\begin{lstlisting}
from hpo import Arch
arch = Arch([0, 1, 2, 1], ['gcn', 'gin', 'fc', 'cheb']) 
\end{lstlisting}
\end{footnotesize}

We assume all leaf nodes (i.e., nodes without descendants) are connected to the output, so there is no need to specific the output node. Besides, the list can be specified in any order, e.g., the following code can specific the same architecture:
\begin{footnotesize}
\begin{lstlisting}
arch = Arch([0, 1, 1, 2], ['gcn', 'cheb', 'gin', 'fc']) 
\end{lstlisting}
\end{footnotesize}

Then, four recorded metrics in the benchmark including the validation and test performance, the latency, and the number of parameters, can be obtained by a look-up table: 
\begin{lstlisting}
info = bench[arch.valid_hash()]
info['valid_perf']  # validation performance
info['perf']        # test performance
info['latency']     # latency
info['para']        # number of parameters
\end{lstlisting}

We provide the full data, including the training/validation/testing performance at each epoch at: 
\url{https://figshare.com/articles/dataset/NAS-bench-Graph/20070371}.
Since we run each dataset with three random seeds, each dataset has 3 files. 
The full metric can be obtained similarly as follows:

\begin{lstlisting}
from readbench import read
bench = read('cora0.bench')  # dataset and seed
info = bench[arch.valid_hash()]
epoch = 50
info['dur'][epoch][0]   # training performance
info['dur'][epoch][1]   # validation performance
info['dur'][epoch][2]   # testing performance
info['dur'][epoch][3]   # training loss
info['dur'][epoch][4]   # validation loss
info['dur'][epoch][5]   # testing loss
info['dur'][epoch][6]   # best performance
\end{lstlisting}

We have also provided the source codes of using our benchmark together with two public libraries for GraphNAS, AutoGL and NNI. See \url{https://github.com/THUMNLab/AutoGL/tree/agnn} and \url{https://github.com/THUMNLab/NAS-Bench-Graph/blob/main/runnni.py} for details.

\begin{figure*}[!ht]
  \hspace{-20pt}
  \centering
  \subfigure[Cora]{
  \includegraphics[width=0.33\textwidth]{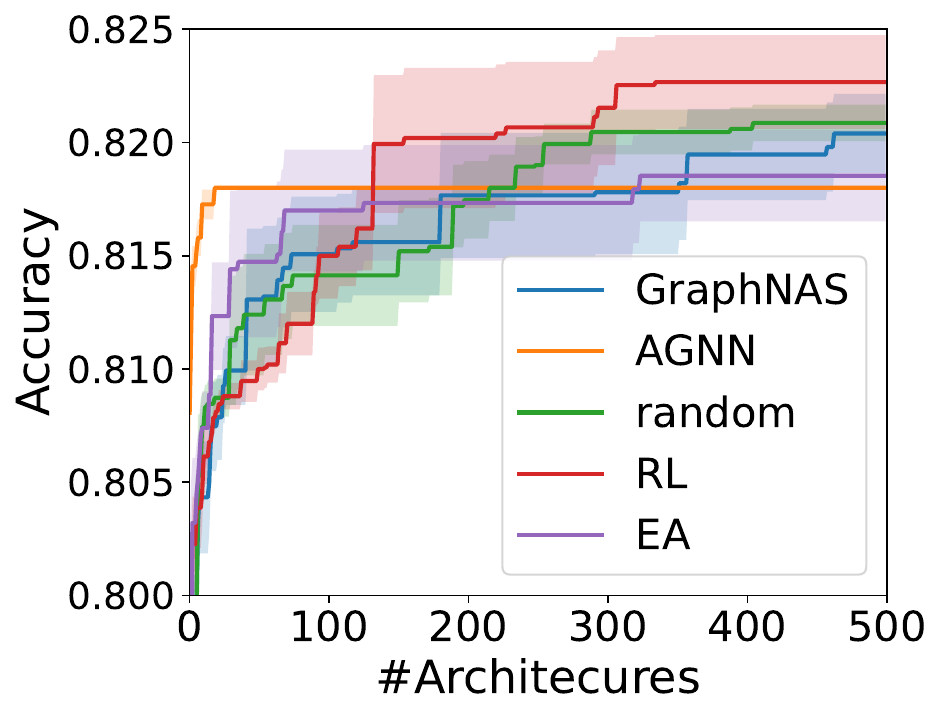}}
  \subfigure[CiteSeer]{
  \includegraphics[width=0.33\textwidth]{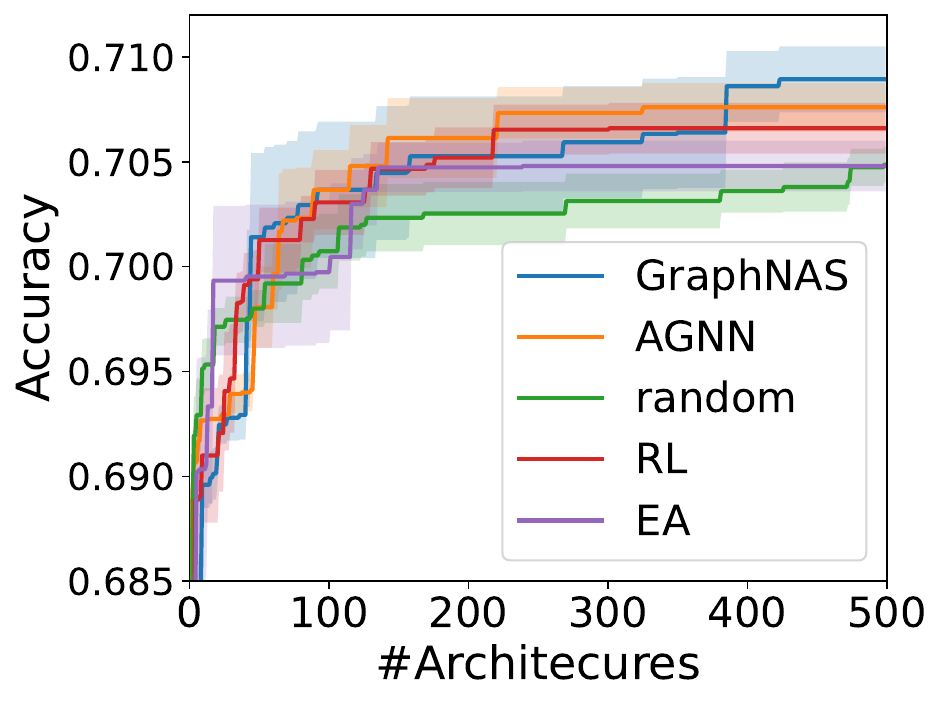}}
  \subfigure[PubMed]{
  \includegraphics[width=0.33\textwidth]{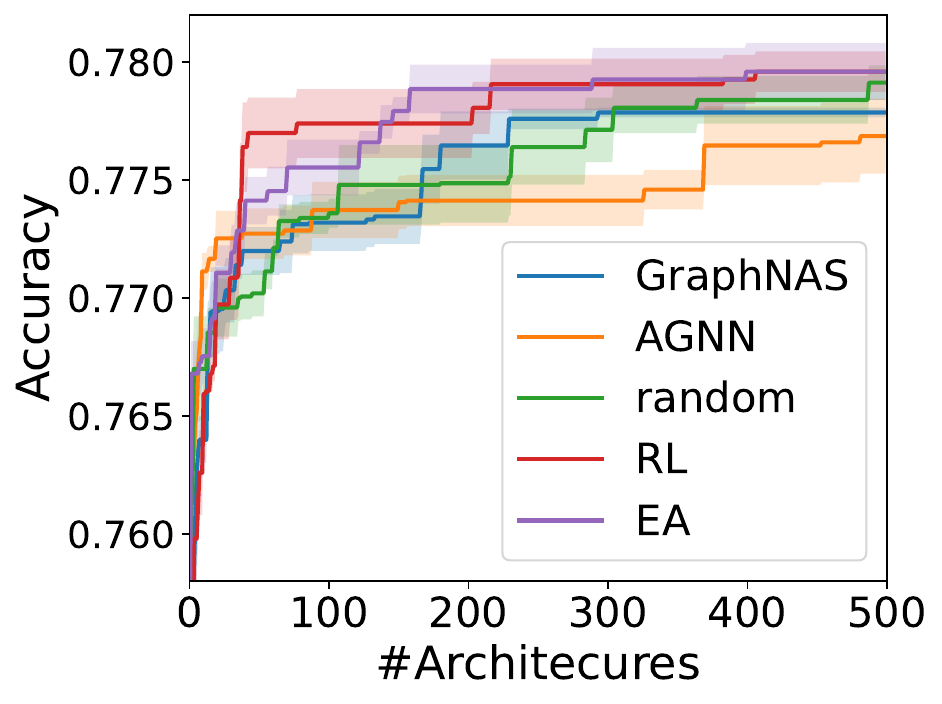}}
  
  \hspace{-20pt}
  \subfigure[Coauthor-CS]{
  \includegraphics[width=0.33\textwidth]{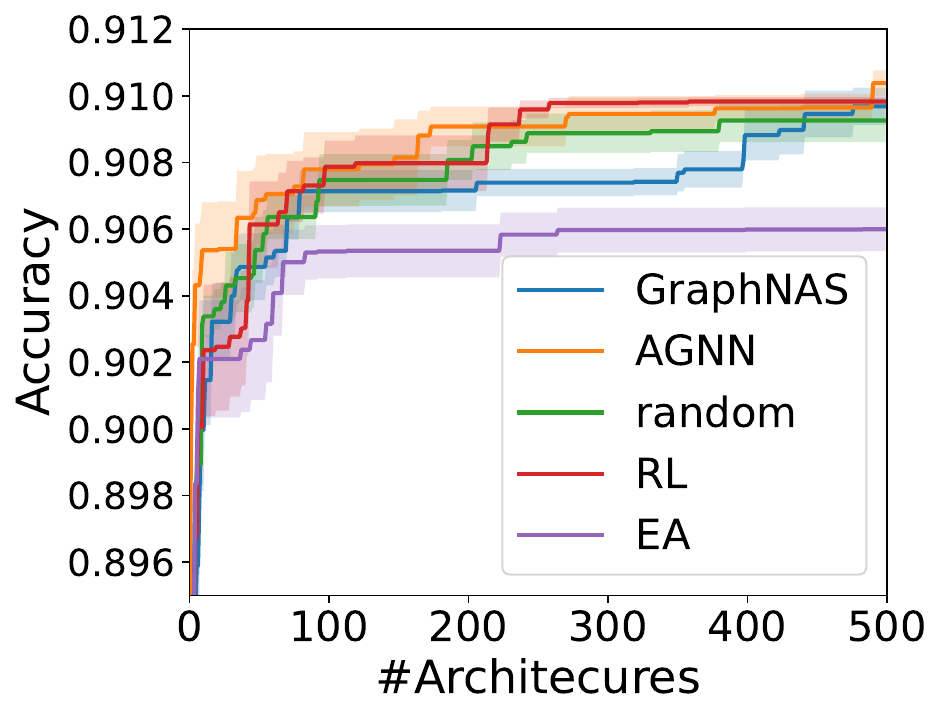}}
  \subfigure[Coauthor-Physics]{
  \includegraphics[width=0.33\textwidth]{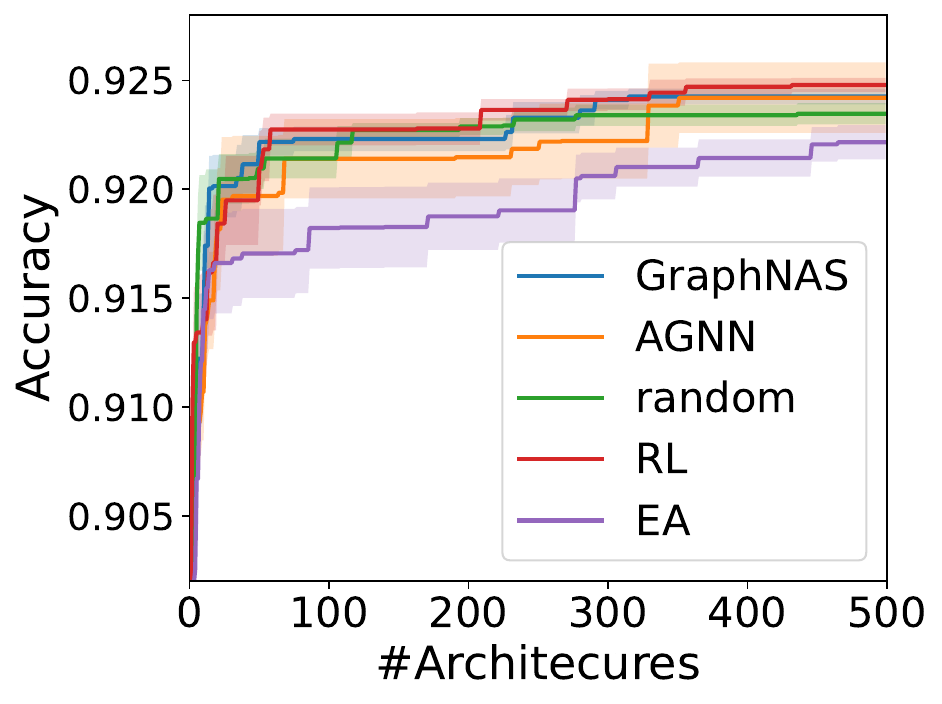}}
  \subfigure[Amazon-Photo]{
  \includegraphics[width=0.33\textwidth]{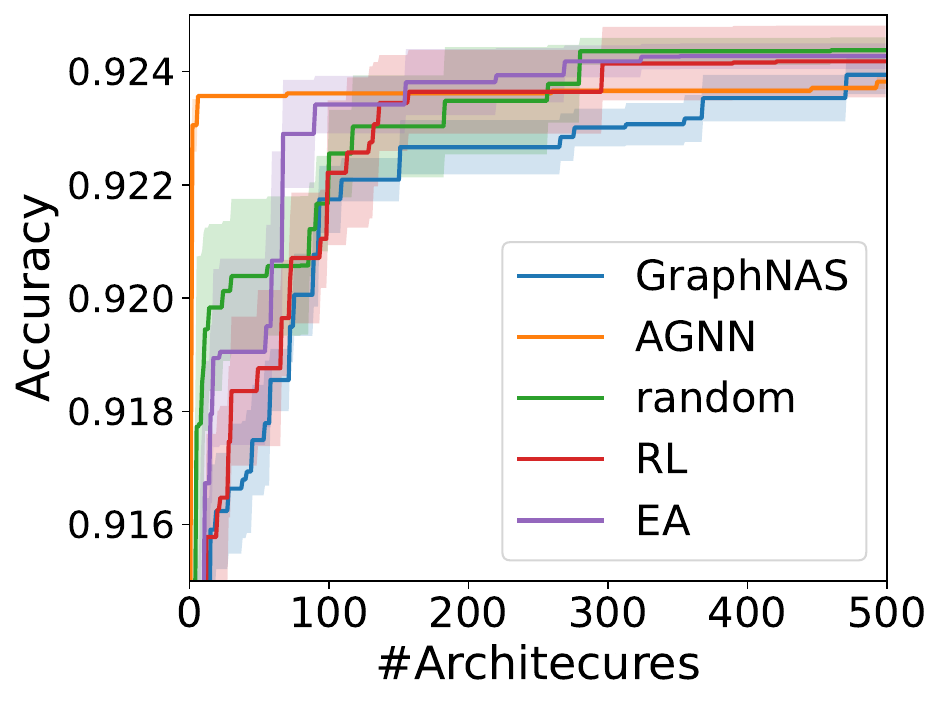}}
  
  \hspace{-20pt}
  \subfigure[Amazon-Computers]{
  \includegraphics[width=0.33\textwidth]{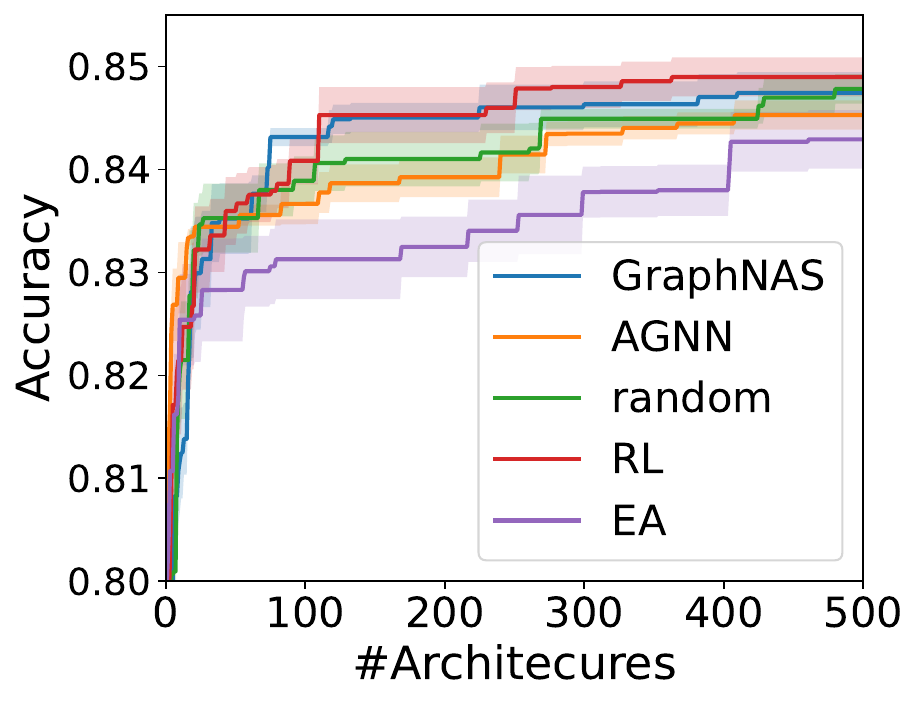}}
  \subfigure[ogbn-arXiv]{
  \includegraphics[width=0.33\textwidth]{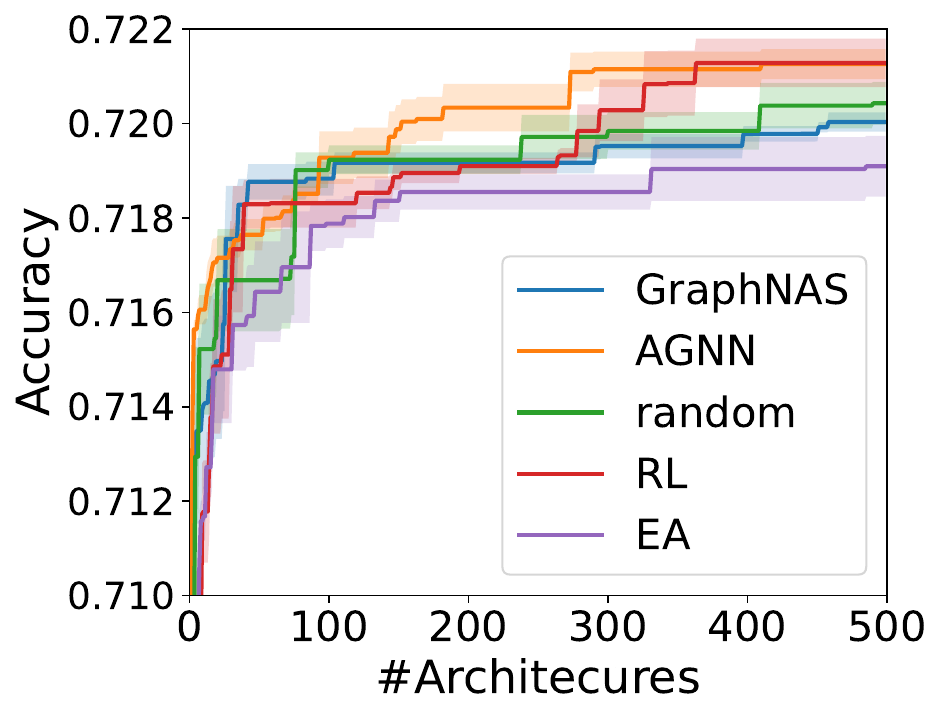}}
  \subfigure[ogbn-proteins]{
  \includegraphics[width=0.33\textwidth]{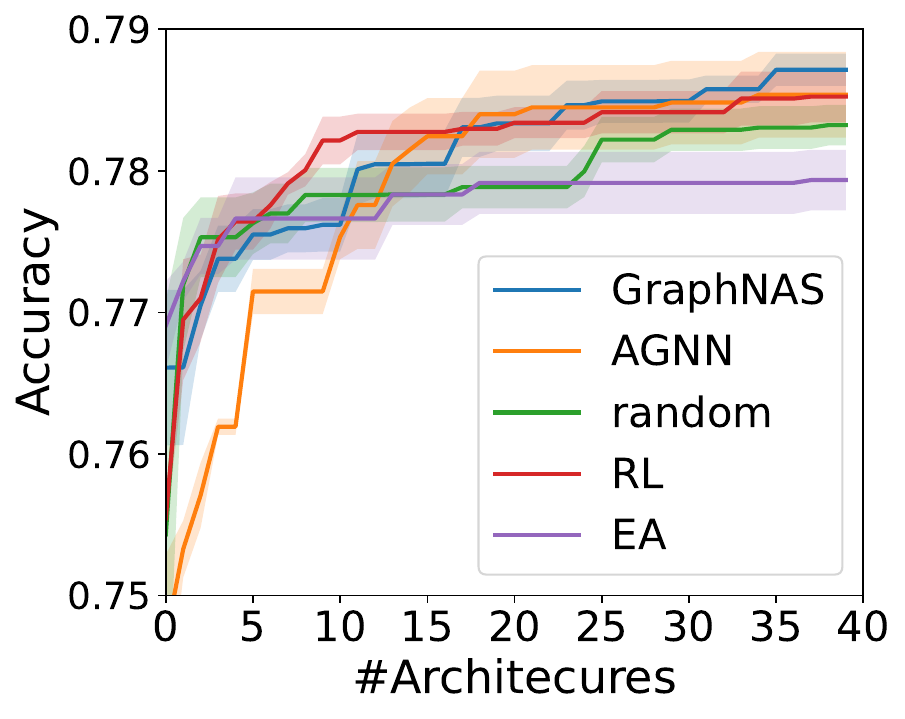}}
  \caption{
    The learning curve depicting the optimal performance as a function of the number of searched architectures is presented herein. The reported results are obtained by averaging measurements from five independent experiments, each conducted with distinct random seeds. The background of the figure displays the standard errors associated with the reported values.
    }
  \label{fig:curve} 
  \end{figure*}

From the experimental findings, it is evident that all algorithms exhibit superior performance compared to the top 5\% benchmark, indicating their ability to acquire meaningful patterns in the NAS-Bench-Graph. Nevertheless, no algorithm demonstrates consistent superiority across all datasets.
Notably, Random Search remains a robust baseline in comparison to alternative methods and even achieves the highest performance on two datasets, partially supporting the findings of Li \textit{et al.}~\cite{li2020random} in the context of general NAS.
These results underscore the pressing need for further research in the domain of graph NAS.

In order to delve into the learning behaviors of diverse graph NAS methods, we present the curves depicting optimal performance as a function of the number of architectures, as illustrated in Figure~\ref{fig:curve}. It is evident that different algorithms exhibit distinct characteristics. For instance, EA and AGNN demonstrate sporadic "jumps" in performance, indicating substantial performance improvements, while RL exhibits more gradual and consistent performance enhancements. A meticulous examination of these learning curves may serve as a source of inspiration for the development of novel algorithms for graph NAS.

\section{Future Directions}\label{sec:future}
We have discussed existing literature in automated graph machine learning approaches and libraries. Our discussion in detail contains how HPO and NAS can be applied to graph machine learning to handle problems in automated graph machine learning. We also introduce AutoGL, a dedicated framework and library for automated graph machine learning. 
In this section, we will suggest future directions deserving further investigations from both academia and industry.
There exist plenty of challenges and opportunities worthy of future explorations. 

\begin{itemize}
\item \textbf{Scalability}: AutoML has been successfully applied to various graph scenarios, however, there are still lots of future directions deserving further investigation regarding scalability to large-scale graphs.
On the one hand, although HPO for large-scale graph machine learning has been preliminarily explored in literature~\cite{tu2019autone}, the Bayesian Optimization utilized in the model suffers from limited efficiency. Thus it will be interesting and challenging to explore how we can reduce the computational costs to realize fast hyper-parameter optimization.
On the other hand, the scalability of NAS for graph machine learning has drawn few attentions from the researchers despite applications involving large-scale graphs are very common in real world, 
leaving a large space for further explorations. 

\item \textbf{Explainability}: Existing automated graph machine learning approaches are mainly based on black-box optimizations. For example, it is unclear why certain NAS models can perform better compared with others, and the explainability of NAS algorithms still lack systematic research efforts. There have been some preliminary studies on explainability of graph machine learning~\cite{yuan2020explainability}, and on explainable graph hyper-parameter optimization~\cite{wang2021explainable} via hyper-parameter importance decorrelation. However, further and deeper investigations on the explainability of automated graph machine learning are still of great importance.

\item \textbf{Out-of-distribution generalization}: When applied to new graph datasets and tasks, there still need huge human efforts to construct task-specific graph HPO configurations and graph NAS frameworks, e.g., spaces and algorithms. The generalization of current graph HPO configurations and NAS frameworks are limited, especially training and testing data come from different distributions~\cite{li2021oodgnn,li2022out,li2022learning,li2023invariant}. It will be a promising direction to study the out-of-distribution generalization abilities for both graph HPO and graph NAS algorithms which are capable of handling continuously and rapidly changing tasks.

\item \textbf{Robustness}: Since many applications of AutoML on graphs are risk-sensitive, e.g., finance and healthcare, the robustness of the models is indispensable for actual usages. Though there exist some initial studies on the robustness~\cite{sun2018adversarial} of graph machine learning, how to generalize these techniques into automated graph machine learning has not been explored.

\item \textbf{Graph models for AutoML}: In this paper, we mainly focus on 
how AutoML methods are extended to graphs. The other direction, i.e., using graphs to help AutoML, is also feasible and promising. For example, we can model neural networks as a directed acyclic graph (DAG) to analyze their structures~\cite{xie2019exploring,you2020graph} or adopt GNNs to facilitate NAS~\cite{shi2020bridging,zhang2018graph,dudziak2020brp,qin2021GQNAS}. Ultimately, we expect graphs and AutoML to form tighter connections and further facilitate each other.

\item \textbf{Hardware-aware models}: To further improve the scalability of automated graph machine learning, hardware-aware models may be a critical step, especially in real industrial environments. Both hardware-aware graph models~\cite{auten2020hardware} and hardware-aware AutoML models~\cite{cai2018proxylessnas,tan2019mnasnet,jiang2020hardware} have been studied, but integrating these techniques is still in the early stage and poses significant challenges.

\item \textbf{Comprehensive evaluation protocols}: Currently, most AutoML on graphs are tested on small traditional benchmarks such as three citations graphs, i.e., Cora, CiteSeer, and PubMed~\cite{sen2008collective}. However, these benchmarks have been identified as insufficient to compare different graph machine learning models~\cite{shchur2018pitfalls}, not to mention AutoML on graphs. More comprehensive evaluation protocols are needed, e.g., on recently proposed graph machine learning benchmarks~\cite{hu2020open,dwivedi2020benchmarking}, or new dedicated graph AutoML benchmarks~\cite{qin2022bench} similar to the NAS-bench series~\cite{ying2019bench} are needed.

\item \textbf{Broader Scope of Applications}:  While automated graph machine learning techniques have been applied to a range of practical use-cases, there's considerable potential for using these newly-developed techniques in the information retrieval (such as search engines, recommender systems) for achieving effective, reliable, and user-friendly predictions. A viable approach could involve using this specialized domain knowledge as a form of prior to guide both the hyperparameter optimization and  architecture search strategies.

\end{itemize}

\section{Conclusion}\label{sec:conclusion}
In this paper, we discuss the current state-of-the-art automated graph machine learning approaches, libraries. 
In particular, we in depth elaborate how graph hyperparameter optimization (HPO) and graph neural architecture search (NAS) have been developed to facilitate automated graph machine learning. 
We also introduce AutoGL, our dedicated framework and open source library for automated graph machine learning, and NAS-Bench-Graph, our tailored benchmark that enables fair, fully reproducible, and efficient empirical comparisons.
Last but not least, we point out challenges and suggest promising directions deserving further investigations.

\section*{Acknowledgment}
This work is supported by the National Key Research and Development Program of China No.2020AAA0106300 and National Natural Science Foundation of China (No. 62222209, 62250008, 62102222), BNRist under Grant No. BNR2023RC01003, BNR2023TD03006, 
and Beijing Key Lab of Networked Multimedia.

\bibliographystyle{IEEEtran}
\bibliography{autogr-ref}

\normalsize


\begin{IEEEbiography}[{\includegraphics[width=1in,height=1.25in,clip,keepaspectratio]{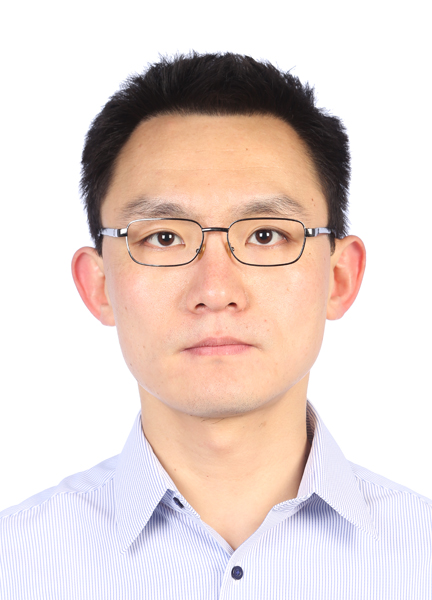}}]{Xin Wang}
 is currently an Associate Professor at the Department of Computer Science and Technology, Tsinghua University. He got both of his Ph.D. and B.E degrees in Computer Science and Technology from Zhejiang University, China. He also holds a Ph.D. degree in Computing Science from Simon Fraser University, Canada. His research interests include multimedia intelligence, machine learning and its applications. He has published over 150 high-quality research papers in ICML, NeurIPS, IEEE TPAMI, IEEE TKDE, ACM KDD, WWW, ACM SIGIR, ACM Multimedia etc., winning three best paper awards including ACM Multimedia Asia. He is the recipient of ACM China Rising Star Award, IEEE TCMC Rising Star Award and DAMO Academy Young Fellow.
\end{IEEEbiography}

\begin{IEEEbiography}[{\includegraphics[width=1in,height=1.25in,clip,keepaspectratio]{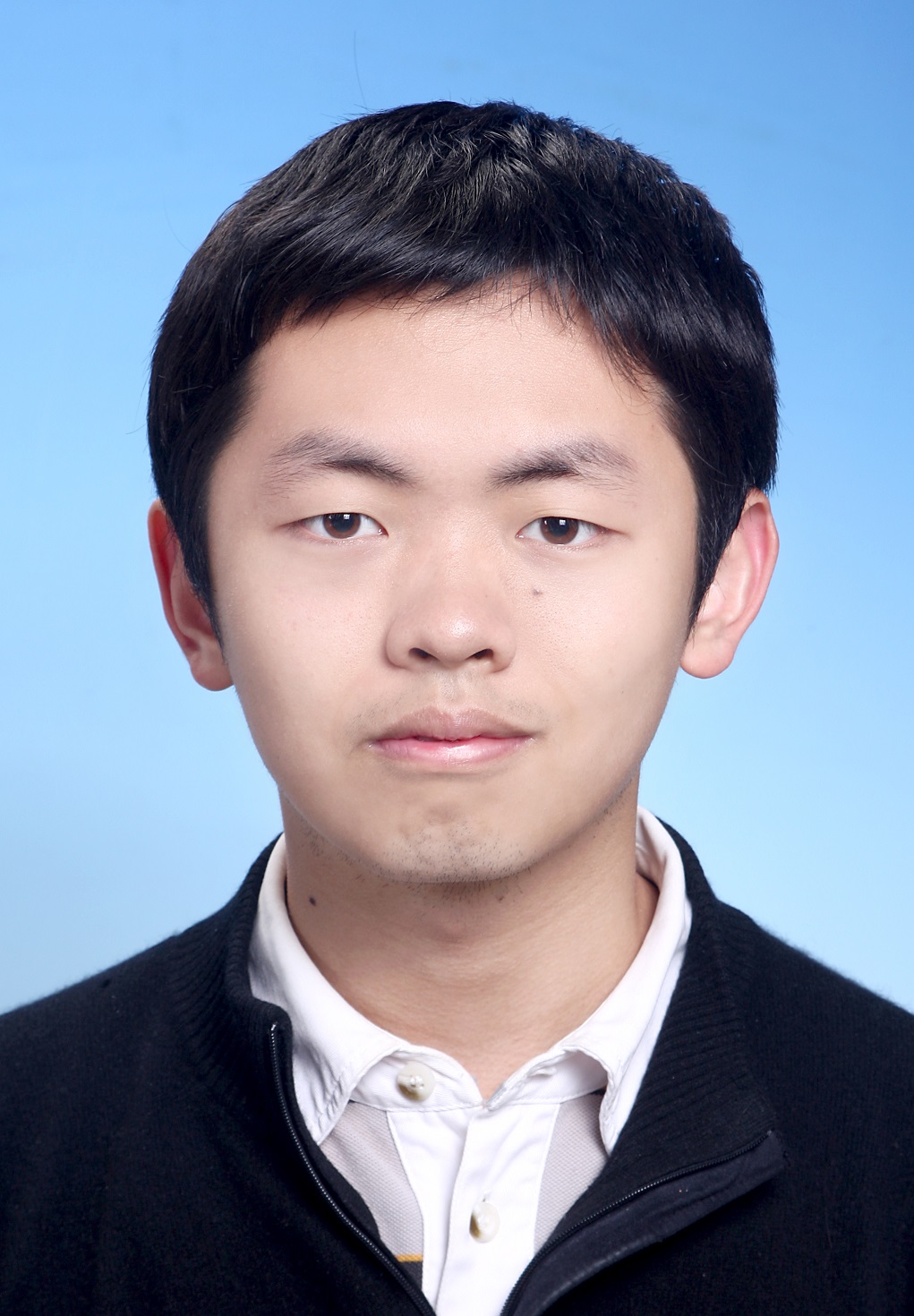}}]{Ziwei Zhang} received his Ph.D. from the Department of Computer Science and Technology, Tsinghua University, in 2021. He is currently a postdoc researcher in the Department of Computer Science and Technology at Tsinghua University. His research interests focus on machine learning on graphs, including graph neural network (GNN) and network embedding (a.k.a. network representation learning). He has published over a dozen papers in prestigious conferences and journals, including KDD, AAAI, IJCAI, and TKDE.
\end{IEEEbiography}

\begin{IEEEbiography}[{\includegraphics[width=1in,height=1.25in,clip,keepaspectratio]{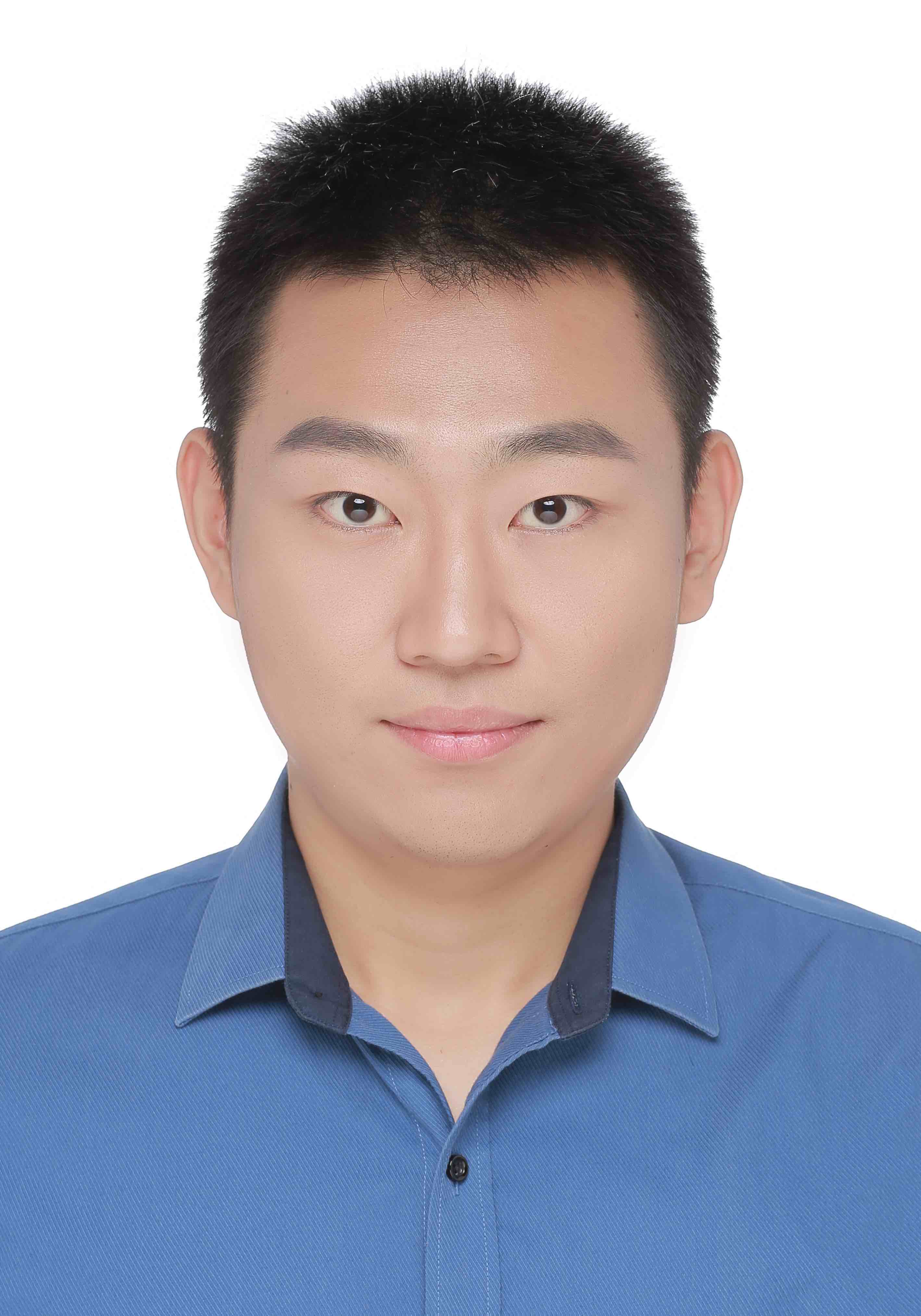}}]{Haoyang Li}
is currently a postdoc researcher at Weill Cornell Medicine of Cornell University. He received his Ph.D. from the Department of Computer Science and Technology of Tsinghua University in 2023. He received his B.E. from the Department of Computer Science and Technology of Tsinghua University in 2018. His research interests are mainly in machine learning on graphs and out-of-distribution generalization. He has published high-quality papers in prestigious journals and conferences, e.g., IEEE TKDE, ACM TOIS, NeurIPS, ICLR, ACM KDD, ACM Web Conference, AAAI, IJCAI, ACM Multimedia, IEEE ICDE, IEEE ICDM, etc., winning one best paper award.
\end{IEEEbiography}

\begin{IEEEbiography}[{\includegraphics[width=1in,height=1.25in,clip,keepaspectratio]{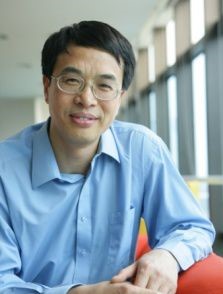}}]{Wenwu Zhu}
	is currently a Professor in the Department of Computer Science and Technology at Tsinghua University, the Vice Dean of National Research Center for Information Science and Technology. Prior to his current post, he was a Senior Researcher and Research Manager at Microsoft Research Asia. He was the Chief Scientist and Director at Intel Research China from 2004 to 2008. He worked at Bell Labs New Jersey as Member of Technical Staff during 1996-1999. He received his Ph.D. degree from New York University in 1996.
	
	His current research interests are in the area of data-driven multimedia networking and multimedia intelligence. He has published over 350 referred papers, and is inventor or co-inventor of over 50 patents. He received eight Best Paper Awards, including ACM Multimedia 2012 and IEEE Transactions on Circuits and Systems for Video Technology in 2001 and 2019.  
	
	He currently serves EiC for IEEE Transactions for Circuits and Systems for Video technology. He served as EiC for IEEE Transactions on Multimedia (2017-2019) and the chair of the steering committee for IEEE Transactions on Multimedia (2020-2022). He serves as General Co-Chair for ACM Multimedia 2018 and ACM CIKM 2019, respectively. He is an AAAS Fellow, IEEE Fellow, SPIE Fellow, and a member of The Academy of Europe (Academia Europaea).
\end{IEEEbiography}




\end{document}